\documentclass[lettersize,journal]{IEEEtran}

\usepackage{mydef}
\usepackage{cite}
\usepackage{amsmath,amssymb,amsfonts,bm}
\usepackage{algorithmic}
\usepackage{graphicx}
\usepackage{balance}
\usepackage{enumitem}
\usepackage{textcomp}
\usepackage{bbding} 
\usepackage{xcolor}
\usepackage{threeparttable}
\usepackage{algorithm}
\usepackage{multirow}
\usepackage{url}
\usepackage{booktabs}
\usepackage{array}
\usepackage{caption}
\usepackage{colortbl}
\usepackage{balance}
\usepackage{float,color}
\usepackage{romannum}

\def\BibTeX{{\rm B\kern-.05em{\sc i\kern-.025em b}\kern-.08em
    T\kern-.1667em\lower.7ex\hbox{E}\kern-.125emX}}

\def \ca {{\mathcal A}}

 \def \cs {{\mathcal S}}

\begin{document}
\def\method{}

%
\title{Graph Neural Networks in Intelligent Transportation Systems: Advances, Applications and Trends}
%
%
%
%

\author{Hourun Li, Yusheng Zhao, Zhengyang Mao, Yifang Qin, Zhiping Xiao, Jiaqi Feng, Yiyang Gu, \\Wei Ju, \IEEEmembership{Member, IEEE}, Xiao Luo, and Ming Zhang
\IEEEcompsocitemizethanks{
\IEEEcompsocthanksitem Hourun Li, Yusheng Zhao, Zhengyang Mao, Yifang Qin, Jiaqi Feng, Yiyang Gu, Wei Ju, and Ming Zhang are with School of Computer Science, National Key Laboratory for Multimedia Information Processing, Peking University-Anker Embodied AI Lab, Peking University, Beijing, China. (e-mail: lihourun@stu.pku.edu.cn, yusheng.zhao@stu.pku.edu.cn, zhengyang.mao@stu.pku.edu.cn, qinyifang@pku.edu.cn, sf77@stu.pku.edu.cn, yiyanggu@pku.edu.cn, juwei@pku.edu.cn, mzhang\_cs@pku.edu.cn)
\IEEEcompsocthanksitem Zhiping Xiao and Xiao Luo are with Department of Computer Science, University of California, Los Angeles, USA. (e-mail: patricia.xiao@cs.ucla.edu, xiaoluo@cs.ucla.edu)

}
}

\markboth{Journal of \LaTeX\ Class Files,~Vol.~14, No.~8, August~2021}%
{Shell \MakeLowercase{\textit{et al.}}: A Sample Article Using IEEEtran.cls for IEEE Journals}

\maketitle

\begin{abstract} 
Intelligent Transportation System (ITS) is crucial for improving traffic congestion, reducing accidents, optimizing urban planning, and more. 
However, the complexity of traffic networks has rendered traditional machine learning and statistical methods less effective. 
With the advent of artificial intelligence, deep learning frameworks have achieved remarkable progress across various fields and are now considered highly effective in many areas. 
Since 2019, Graph Neural Networks (GNNs) have emerged as a particularly promising deep learning approach within the ITS domain, owing to their robust ability to model graph-structured data and address complex problems. Consequently, there has been increasing scholarly attention to the applications of GNNs in transportation, which have demonstrated excellent performance.
Nevertheless, current research predominantly focuses on traffic forecasting, with other ITS domains, such as autonomous vehicles and demand prediction, receiving less attention. 
This paper aims to review the applications of GNNs across six representative and emerging ITS research areas: traffic forecasting, vehicle control system, traffic signal control, transportation safety, demand prediction, and parking management. 
We have examined a wide range of graph-related studies from 2018 to 2023, summarizing their methodologies, features, and contributions in detailed tables and lists. Additionally, we identify the challenges of applying GNNs in ITS and propose potential future research directions.

\end{abstract}


\section{Introduction} \label{introduction}

\IEEEPARstart{A}{s} cities grow and transportation systems evolve, several issues have become increasingly apparent, such as traffic congestion, environmental pollution, and a rising number of traffic accidents. 
To address these challenges and improve traffic flow, route planning, and transportation safety, the Intelligent Transportation System (ITS) was introduced over five decades ago in the U.S. \cite{wootton1995intelligent}.
Today, ITS applications are integral to everyday life, including Electronic Toll Collection (ETC), Traffic Management Systems (TMS), Global Positioning Systems (GPS), and Commercial Vehicle Operations (CVO). 
ITS encompasses a broad range of areas, including traffic forecasting, autonomous vehicles, traffic signal control, and more.
Notably, traffic forecasting has emerged as a prominent research area, garnering significant attention due to its critical applications in optimizing route planning, facilitating road traffic, and reducing traffic accidents.

According to Verses et al. \cite{veres2019deep}, addressing practical challenges such as managing massive and noisy data, scalability, and generalization remains difficult. 
Over the past three decades, statistical methods, including simple linear time series models like Autoregressive Integrated Moving Average (ARIMA) \cite{lee1999application, williams2001multivariate}, and traditional machine learning methods such as Logistic Regression (LR), Support Vector Regression (SVR), and k-Nearest Neighbors (KNN) \cite{jian2013evaluation, wu2004travel, chang2012dynamic} have been proposed to tackle these issues.
However, the increasing volume of data and the complexity of road conditions have made traditional methods less effective.
Consequently, there is a need for more efficient algorithms and scalable models to fully leverage massive data and develop accurate and efficient ITS solutions. 
Additionally, advancements in computational techniques, such as graphical processing units, have propelled the effectiveness of deep learning models.
Since 2015, deep-learning models for traffic forecasting have seen significant progress, with GNNs emerging as the most popular models after 2019 \cite{fan2020deep}. 
GNNs excel not only at modeling graph-structured problems but also at capturing temporal-spatial dependencies and representing relationships in non-Euclidean spaces \cite{fan2020deep, jiang2022graph, rahmani2023graph}.



After conducting a comprehensive survey of research in the field of ITS, we found that a significant portion of studies primarily focuses on traffic forecasting. 
However, we believe that other critical domains within ITS also warrant greater attention. Despite the recent shift toward promising techniques such as deep learning and reinforcement learning, GNNs still require more exploration and application.
Considering the graph-based nature of traffic networks and the inherent advantages offered by GNNs, we argue that GNNs represent a highly promising and competitive solution for ITS. 
Our investigation centers on papers related to GNNs in ITS published between 2018 and 2023, providing a detailed summary of their contributions. 
We also identified key research challenges within ITS and proposed potential future directions for leveraging GNNs.
Our main contribution can be summarized as follows:

\begin{itemize}
  \item \textit{Comprehensive Review.} Extensive research work and surveys from 2018 to 2023 for ITS are reviewed in detail, covering six distinct research domains rather than focusing solely on traffic forecasting. Moreover, we offer an in-depth analysis of the papers reviewed, summarize methods and challenges, and present informative tables and lists.

  \item \textit{A Comprehensive Taxonomy.} We categorized the reviewed studies based on various criteria, including research domain, graph methods, and domain-specific challenges. This approach helps readers comprehensively understand each research domain in ITS.
  
  \item \textit{Challenges and Future Directions.} After a comprehensive review, we summarize the key challenges when applying GNNs in ITS and propose potential future directions, which provide valuable insights for researchers looking to explore and advance this field.

\end{itemize} 

  

We organize the rest of the survey as follows:
In section \ref{surveys}, we quickly review the related surveys in transportation domains and briefly introduce them.
In Section \ref{Background}, we present the foundational knowledge on ITS and GNNs, along with a discussion on problem formulation.
In Section \ref{GNNs}, we investigate and review extensive graph-based studies in six domains, including traffic forecasting, vehicle control system, traffic signal control, transportation safety, demand prediction, and parking management. 
In Section \ref{challenges}, we summarize the challenges and potential future directions in GNNs for ITS based on the previous review results.
Finally, we present the conclusions in Section \ref{conclusion}.
\section{Related Surveys} \label{surveys}
\begin{figure*}[htp]
    \centering
    \includegraphics[width=1.0\textwidth]{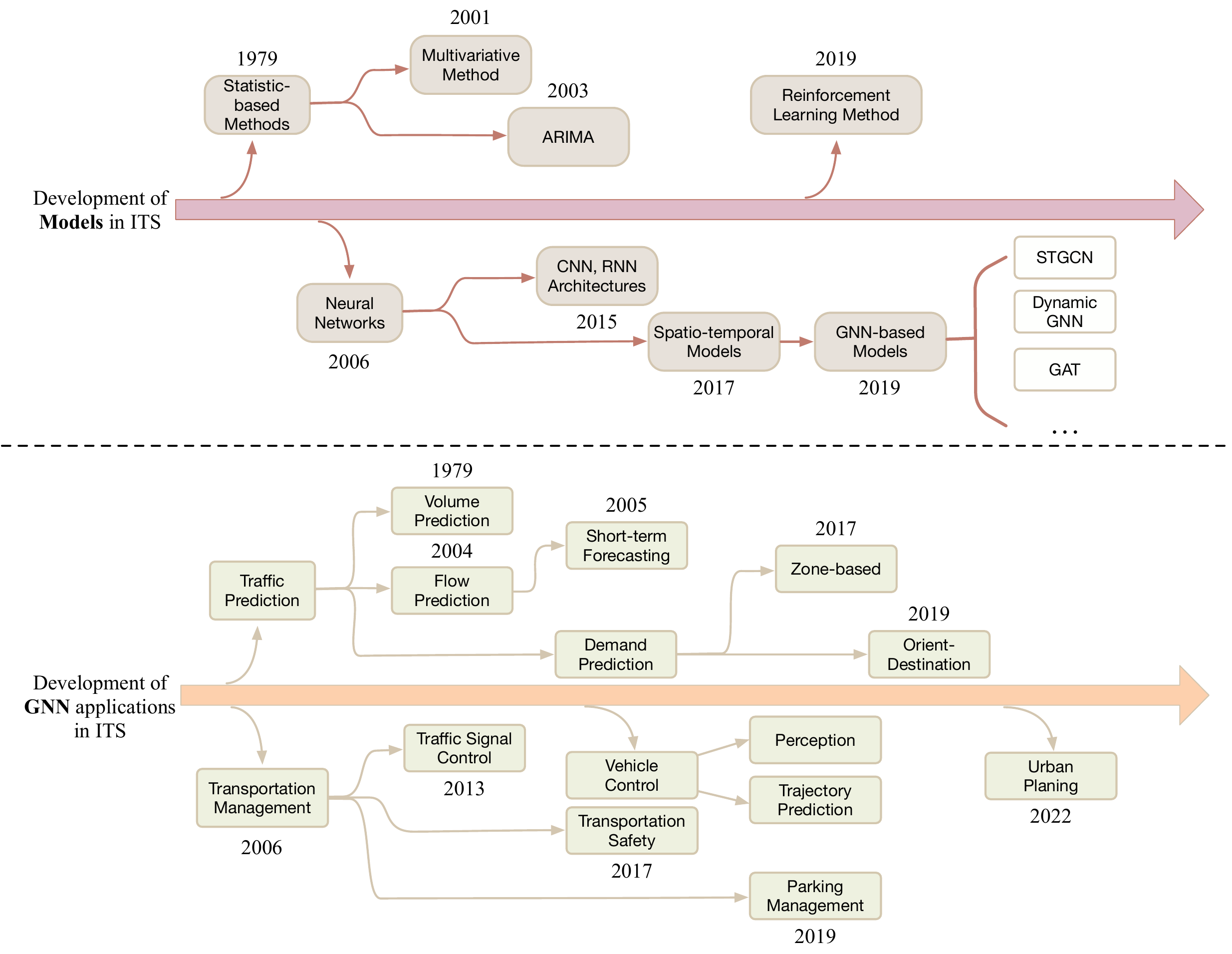}
    \caption{Development of Models and Application in Intelligent Transportation System}
    \label{ITS_development}
    \vspace{-0.2in}
\end{figure*}
In this section, we review the most relevant and representative surveys in Intelligent Transportation Systems (ITS), primarily focusing on those published in the past five years. 
Special emphasis is given to approaches based on Graph Neural Networks (GNNs), for which we provide a thorough introduction.


As previously mentioned, ITS encompass a variety of research fields. 
However, recent surveys have predominantly concentrated on traffic forecasting \cite{cui2019traffic, tedjopurnomo2020survey, fan2020deep, yin2021deep, jiang2022graph, jiang2023graph, vlahogianni2014short, lee2021short}, with only a few exploring other ITS domains \cite{rahmani2023graph, haghighat2020applications, guerrero2021deep, veres2019deep}.
Traffic flow prediction has been the most prominent research topic in traffic forecasting since 2015, as highlighted by Liu et al. \cite{liu2021scientometric}. 
Surveys prior to 2015 primarily focused on statistical methods \cite{ahmed1979analysis, levin1980forecasting, lee1999application, williams2001multivariate, williams2003modeling, kamarianakis2003forecasting} and traditional machine learning models \cite{jian2013evaluation, wu2004travel, chang2012dynamic, vanajakshi2004comparison}. 
Notable early surveys by Vlahogianni et al. \cite{vlahogianni2004short, vlahogianni2014short} in 2004 and 2014 concentrated on short-term traffic forecasting. 
However, these traditional methods have limitations in addressing complex transportation problems due to their shallow architectures. 
With advancements in theory and hardware, deep learning models have gained popularity since the mid-2010s, significantly advancing traffic forecasting \cite{tedjopurnomo2020survey, fan2020deep}. 
Since 2019, GNNs have become particularly prominent, underscoring their growing importance in ITS \cite{jiang2022graph, jiang2023graph, cui2019traffic, rahmani2023graph}.
In recent years, researchers have increasingly focused on the temporal and spatial dependencies of traffic data \cite{bui2022spatial, li2023graph}, leading to the exploration of new research trends and directions. The detailed development of models and GNN applications in ITS is illustrated in Fig.\ref{ITS_development}.

It is worth noting that Jiang et al. published a comprehensive review of GNNs for traffic forecasting \cite{jiang2022graph}, which summarizes the research on this topic. 
They reviewed 212 articles published between 2018 and 2020, provided a detailed taxonomy of problems and methods, and compiled information on open-source data and code resources.
In the following year, they extended their work with another survey \cite{jiang2023graph}, which updates the previous review by describing the latest research developments and trends up to 2022. 
This follow-up also highlighted specific challenges and proposed informative future directions. 
While their surveys offer an exhaustive overview of traffic forecasting, they do not cover other research areas within ITS.



\textbf{The most relevant survey} is the work by Rahmani et al. \cite{rahmani2023graph}, which stands out as the most recent and comprehensive review of GNNs in general ITS research. This review covers fundamental concepts such as graphs and GNNs and spans seven ITS research domains: traffic forecasting, demand prediction, autonomous vehicles, intersection management, parking management, urban planning, and transportation safety. 
For each domain, it provides a brief introduction followed by an independent review of the relevant literature.
In contrast, our survey offers a deeper examination of the application of graphs and GNNs in ITS. 
We provide an extensive exploration of graph construction, GNN customization for specific challenges, and performance assessment across various ITS domains, offering insights that are not covered in \cite{rahmani2023graph}. 
Furthermore, instead of introducing individual papers independently, our systematic approach to summarizing related literature offers readers a more reliable and comprehensive analysis. 
Our review encompasses six domains: traffic forecasting, autonomous vehicles, traffic signal control, transportation safety, demand prediction, and parking management. It offers a more thorough and detailed exploration of ITS compared to existing surveys.

\section{Background} \label{Background}

In this section, we will provide background information on ITS, graphs, and GNNs. 
We will begin by introducing the key concepts of ITS and the associated research domains. 
Next, we will explain fundamental graph concepts, covering different types of graph data and their characteristics.
Finally, we will provide an overview of GNN variants and basic GNN models, laying the foundation for the more detailed discussions in the subsequent sections.

\subsection{Intelligent Transportation System (ITS)}



\begin{figure*}[!t]
    \centering
    \includegraphics[width=0.95\textwidth]{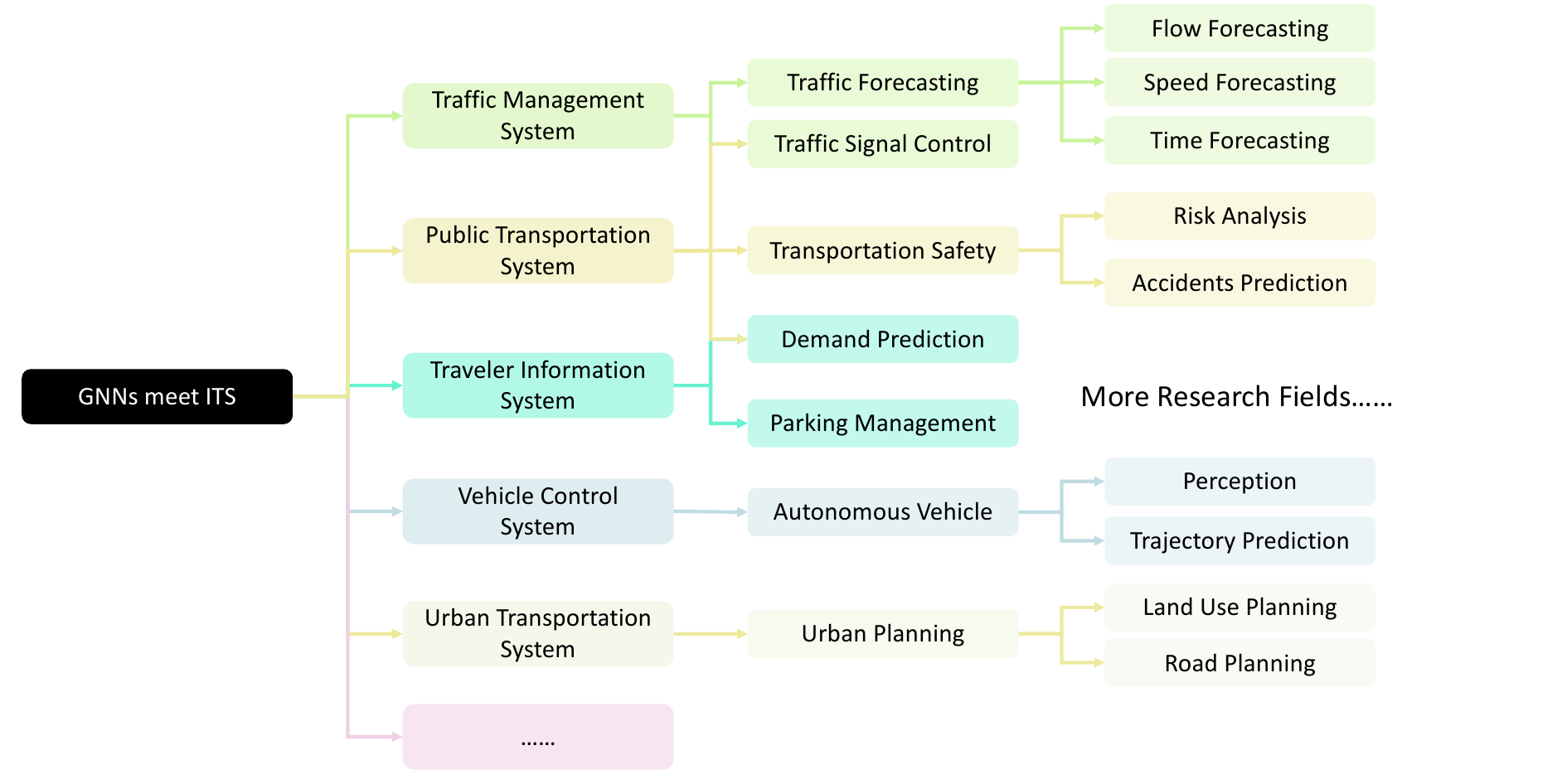}
    \caption{Research Fields in Intelligent Transportation System}
    \label{ITS_domains}
    \vspace{-0.2in}
\end{figure*}

Before $1980$, ITS was primarily a forward-looking concept aimed at overcoming the limitations of surface transportation capacity~\cite{auer2016history, wootton1995intelligent}. 
During this period, the focus was on improving road network efficiency through optimized traffic signals and in-vehicle navigation systems~\cite{mundy1981management, bushnell1981transportation, auer2016history}. 
In the 21st century, technological advancements have made traffic data more accessible, leading to a surge in data-driven approaches~\cite{zhang2011data}. 
Real-time traffic data is now used extensively for traffic management, including road condition prediction, congestion identification, and navigation~\cite{zhang2011data, jia2017data, gunther2017big, saroj2018smart, wang2018data, njoku2023prospects}. 
Nowadays, ITS is a rapidly growing interdisciplinary field~\cite{lin2017intelligent}, offering innovative services to enhance performance, improve travel safety, and provide useful information to users.
Zhang et al.~\cite{zhang2011data} identify six primary subsystems in ITS. 
For our focus on applying GNNs to address specific issues, we have further categorized these subsystems into specific research fields: traffic forecasting, vehicle control system, traffic signal control, transportation safety, demand prediction, and parking management, as illustrated in Fig.\ref{ITS_domains}.

\subsection{Graphs and Graph Neural Networks in ITS}
In this section, we will briefly introduce graphs and GNNs, which are necessary for the following discussion.
\subsubsection{Graphs and Graph Types} \ 
In general, a graph consists of a set of nodes $\mathcal{V} = \{ v_1, v_2, \dots, v_{|\mathcal{V}|} \}$ and a corresponding set of edges $\mathcal{E}$, denoted as $\mathcal{G} = ( \mathcal{V}, \mathcal{E} )$. 
The graph $\mathcal{G}$ can be represented by an adjacency matrix $\mathbf{A} \in \mathbb{R}^{|\mathcal{V}| \times |\mathcal{V}|}$, where $\mathbf{A}_{ij}=1$ if $e_{ij}=(v_i, v_j)\in\mathcal{E}$, and $\mathbf{A}_{ij}=0$ otherwise. 
In addition, each node $v_i$ can be associated with node features $x_i \in \mathbb{R}^d$, and the feature matrix of the graph is denoted as $\mathbf{X} \in \mathbb{R}^{|\mathcal{V}| \times d}$, where $d$ is the dimension of the features. 
Therefore, a graph can also be represented as $\mathcal{G}=\{\mathbf{X}, \mathbf{A}\}$ and the graph can be further categorized into several types:

\begin{itemize}
    \item \textbf{Directed/Undirected Graphs}. In an undirected graph, if there is an edge $(v_i, v_j) \in \mathcal{E}$, then $(v_j, v_i) \in \mathcal{E}$ must also hold, and vice versa. While in a directed graph, this constraint does not apply.
    
    \item \textbf{Weighted/Unweighted Graphs}. In a weighted graph, edges are assigned weight values to indicate their varying importance or other necessary information. In an unweighted graph, all edges are considered equal.
    
    \item \textbf{Signed/Unsighed Graphs}.Signed graphs consist of edges with positive or negative signs, representing different types of relationships. In contrast, unsigned graphs have edges without sign distinctions, indicating neutral relationships.
    
    \item \textbf{Static/Dynamic Graphs}. Static graphs have fixed node and edge features that remain constant throughout. 
    In contrast, dynamic graphs evolve, where new nodes or edges can emerge or disappear at any time step.Therefore, it is essential to model temporal information accurately to capture the changes over time.Normally, we represent a dynamic graph as a sequence of static graph screen-shots:
    \begin{displaymath}
        \mathcal{G} = \{ \mathcal{G}_1, \mathcal{G}_2, \dots \mathcal{G}_T \}\,,
    \end{displaymath}
    where $\mathcal{G}_t = (\mathcal{V}_t, \mathcal{E}_t), t \in \{1, 2, \dots, T\}$ and $T$ is the total number of time steps.
    
    \item \textbf{Homogeneous/Heterogeneous Graphs}. Homogeneous graphs consist of nodes and edges of a single type, representing uniform entities and relationships. In contrast, heterogeneous graphs involve multiple types of nodes and edges, capturing diverse entities and complex interactions.
    
    \item \textbf{Nested Graphs}. Nested graphs are hierarchical, with nodes that can be expanded into subgraphs containing additional nodes and links. Nested graphs are valuable in ITS for managing multi-level information flows and conducting multi-level prediction tasks.
    
    \item \textbf{Hypergraphs}. Hypergraphs enhance traditional graphs by allowing edges to connect any number of nodes, which enables the modeling of more complex relationships among groups of entities.
\end{itemize}


\subsubsection{Graph Construction in ITS}
Graph data is highly effective for representing traffic flow data. 
For instance, in a system with \( N \) road sections, we can model this as a graph \(\mathcal{G} = (\mathcal{V}, \mathcal{E})\),  where \(\mathcal{V} = \{ v_1, v_2, \dots, v_N \}\) denotes the set of nodes, each corresponding to a road section. 
The edges, represented by \(\mathcal{E}\), capture the relationships between these nodes, such as connectivity. 
If two road sections are directly connected, an edge is included in the edge set, denoted as \((v_i, v_j) \in \mathcal{E}\), where \(v_i\) and \(v_j\) are the connected road sections, and \(\mathcal{E}\) is the set of all such edges.
This graph construction method provides a clear way to model and analyze traffic flow and connectivity.


However, graph construction varies across different ITS research domains and is highly problem-specific. 
For example, in urban traffic management, a dynamic graph is often used where road intersections are represented as nodes and roads as edges, capturing traffic connectivity and flow. 
Edges in this graph might be weighted to reflect parameters such as travel time, congestion levels, or distance. 
While in demand prediction, nodes might represent bus stops or train stations, while edges could indicate routes with attributes such as frequency, capacity, and schedule adherence. 
The diverse methods of graph construction highlight the need for tailored approaches that address the unique requirements and challenges of each application. 
Therefore, in the following sections, we will explore the typical graph construction methods used in each research domain in detail.

\subsection{Graph Neural Networks (GNNs)}
\subsubsection{Definition of GNNs}
\label{subsubsec:backgroud_graph_gnns}

Graph Neural Networks (GNNs) are designed specifically for graph data. 
A central mechanism of GNNs is the message-passing paradigm \cite{gilmer2017neural}, which iteratively aggregates information from neighboring nodes to capture the structural characteristics of the graph. 
Let $h_v^{(l)}$ indicates the embedding of $v$ at layer $l\in\{1,\dots,L\}$, the message-passing mechanism can be denoted as:
\begin{equation}
    h_v^{(l)} = \mathcal{U}^{(l)}\Bigl(h_v^{(l-1)}, \mathcal{A}^{(l)}\Bigl(\bigl\{h_u^{(l-1)}\bigr\}_{u \in \mathcal{N}(v)}\Bigr)\Bigr),
\end{equation}
where $\mathcal{A}^{(l)}$ and $\mathcal{U}^{(l)}$ represent the message aggregating and embedding updating function at layer $l$ respectively, and $\mathcal{N}(v)$ denotes the neighbors of $v$.
After $L$ iterations, we can obtain the node-level representation $h_v^{(l)}$. The graph-level representation $h_{\mathcal{G}}$ can be obtained by aggregating all node representations at layer$L$ with a readout function. 
Formally, it can be defined as:
\begin{equation}
    h_{\mathcal{G}} = \text{READOUT}(\{h_v^{(L)}\}_{v\in\mathcal{V}}),
\end{equation}
where $\text{READOUT}$ could be averaging or other graph-level pooling functions depending on the model ~\cite{lee2019self,ying2018hierarchical,zhang2018end}.

\subsubsection{Variants of GNNs}
In the rapidly evolving field of GNNs, a variety of variants have emerged as crucial tools for effectively utilizing graph-structured data and tackling the unique challenges across different domains. 
The existing literature offers numerous taxonomies to classify these variants based on structural, operational, and application-oriented characteristics.
For the purpose of our survey, which specifically targets ITS, we will not present an exhaustive taxonomy of GNNs. 
Instead, we will introduce a few preliminary GNN variants that are highly relevant to our discussions on ITS applications. 
For those interested in a more thorough exploration of GNN taxonomies and a broader classification of these networks, we recommend referring to the comprehensive works by Wu et al. \cite{wu2020comprehensive}, Zhou et al. \cite{zhou2020graph}, and Zhang et al. \cite{zhang2020deep}.


\begin{itemize}
    \item \textbf{Convolutional GNNs}. 
    The underpinning of Convolutional GNNs is rooted in the concept of expanding traditional convolution operations from Euclidean to non-Euclidean domains.
    Convolutional GNNs can be categorized into spectral and spatial methods. 
    Spectral methods are based on spectral graph theory but suffer from computational inefficiencies and a lack of localization in the spatial domain. 
    Spatial methods directly compute convolutions by aggregating and transforming features from a node's neighborhood, avoiding costly spectral transformations.
    In conclusion, convolutional GNNs adapt convolutional operations to graph-structured data, offering a robust framework for feature learning on graphs.
    
    \item \textbf{Recurrent-Based GNNs}.
    Recurrent-based GNNs combine the characteristics of GNNs and recurrent neural networks (RNNs) to effectively capture dependencies in sequential and graph-based data. By employing a message-passing mechanism and recurrent updates, recurrent-based GNNs can dynamically adapt to changes in the graph structure. This approach provides a flexible and adaptable framework that can effectively handle spatial and temporal complexities in real-world datasets, particularly in dynamic social and traffic networks.


    \item \textbf{Spatial-Temporal GNNs}.
    Spatial-temporal GNNs integrate the spatial structure of a graph using a spatial graph neural network and a time-series processing model such as RNN, LSTM, or Transformer. This integration allows for simultaneous data processing in both the temporal and spatial dimensions. Therefore, spatial-temporal GNNs are commonly employed to analyze graph data with temporal dynamics, particularly for traffic forecasting ~\cite{jiang2022graph}.
    
    \item \textbf{Graph Autoencoders}. Graph autoencoders (GAEs) are unsupervised generative models designed to encode graph-structured data into a low-dimensional representation using encoders such as GCN ~\cite{abadal2021computing}.
    Subsequently, GAEs reconstruct the original graph from this representation. 
    GAEs are commonly utilized for tasks such as node embedding, link prediction, and graph generation. 
    An example of a well-known GAE is the Variational Graph Autoencoder (VGAE)~\cite{kipf2016variational}.

    \item \textbf{Graph Reinforcement Learning}. 
    Graph reinforcement learning (GRL) is an innovative approach that integrates reinforcement learning (RL) with graph-based representations to address challenges associated with graph-structured data where strategic decision-making is desirable. 
    In GRL, agents develop optimal policies by interacting with graph-structured environments, leveraging the connections and relationships between nodes to inform their decisions. GRL is applied to optimize network routing and conduct molecular design on graphs.

    

\end{itemize}

\section{The Applications of GNNs in ITS} \label{GNNs}
This section will review research across six critical domains within ITS: traffic forecasting, vehicle control system, traffic signal control, transportation safety, demand prediction, and parking management. 
For each domain, we define the core problems, evaluate their potential impact, and identify the associated challenges or requirements.
Next, we present a logically organized review of the research, emphasizing key findings and trends rather than discussing each paper independently. 
Finally, we discuss how graphs are constructed within each specific domain and examine the modifications made to standard GNN to address the unique needs and challenges of each area.


\subsection{Traffic Forecasting}
Traffic forecasting, also  referred to as traffic prediction, has been a prominent research topic in ITS over the last few decades \cite{manibardo2021deep}. 
Its objective is to predict the future traffic state on a road network. 
Based on the observed metrics, traffic forecasting can be roughly categorized into two sub-types: flow forecasting and speed/time forecasting.

\begin{table*}[h]
\centering
\caption{A Comprehensive Overview of Most Related Studies for Traffic Forecasting}
\resizebox{\textwidth}{!}{%
\begin{tabular}{cccccccl}
\toprule
\textbf{Model} & \textbf{Article} & \textbf{Year} & \textbf{Task} & \textbf{Graph Construction} & \textbf{Spatial Module} & \textbf{Temporal Module} & \multicolumn{1}{c}{\textbf{Summary}}\\
\midrule
FPTN            & \cite{zhang2023fptn} & 2023 &  Flow  & road network & transformer & transformer & \begin{tabular}[c]{@{}l@{}}FPTN improves traffic forecasting with sensor-based data division, triple types of \\ embeddings, and an efficient Transformer encoder, reducing computational demands. \end{tabular} \\ \hline
DyHSL            & \cite{zhao2023dynamic} & 2023 &  Flow  & learned hypergraph & HGNN & HGNN &\begin{tabular}[c]{@{}l@{}} DyHSL improves traffic forecasting with hypergraphs for dynamics and interactive \\ convolutions for spatio-temporal relations, effective across multiple datasets.\end{tabular} \\ \hline
DSTAGNN         & \cite{lan2022dstagnn} & 2022 &  Flow  & dynamic & GNN & GNN & \begin{tabular}[c]{@{}l@{}} DSTAGNN dynamically models spatial-temporal road network interactions by \\ utilizing enhanced multi-head attention and multi-scale gated convolution.\end{tabular}\\ \hline
Bi-STAT         & \cite{chen2022bidirectional} & 2022 &  Flow  & road network & transformer & transformer & \begin{tabular}[c]{@{}l@{}} Bi-STAT enhances traffic forecasting with adaptive spatial-temporal transformers, \\ handling diverse task complexities and leveraging past data for improved prediction.\end{tabular}\\ \hline
STFGNN          & \cite{li2021spatial}  & 2021 &  Flow  & road network & GNN & GNN & 
\begin{tabular}[c]{@{}l@{}} STFGNN enhances traffic forecasting by fusing data-driven temporal and spatial \\  graphs and employing gated convolutions, effectively handling long sequences.\end{tabular} \\ \hline
AGCRN           & \cite{bai2020AGCRN} & 2020 &  Flow  & generated & GCN & RNN &\begin{tabular}[c]{@{}l@{}} AGCRN enhances prediction by two adaptive modules, focusing on node-specific \\ patterns and automatic inter-dependency learning without pre-defined graphs. \end{tabular}\\ \hline
HGC-RNN         & \cite{yi2020hypergraph} & 2020 &  Flow  & road network & HGNN & RNN &  \begin{tabular}[c]{@{}l@{}} HGC-RNN leverages hypergraph convolution and RNNs for structured time-series \\ sensor network data, capturing complex structural and temporal dependencies. \end{tabular}\\ \hline
STSGCN          & \cite{2020STSGCN}   & 2020 &  Flow  & road network & GCN & GCN & \begin{tabular}[c]{@{}l@{}}STSGCN models localized spatial-temporal correlations and accounts for heterogen-\\eities across different periods, simplifying spatial-temporal network data forecasting. \end{tabular}\\ \hline
ASTGCN          & \cite{2019ASTGCN}   & 2019 &  Flow  & road network & GCN & attention &  \begin{tabular}[c]{@{}l@{}}ASTGCN improve forecasting with a spatial-temporal attention mechanism and \\ convolutions, focusing on dynamic correlations to make more accurate predictions. \end{tabular}\\ \hline
LRGCN           & \cite{li2019predicting} & 2019 &  Flow  & road network & RGCN & RGCN &  \begin{tabular}[c]{@{}l@{}}LRGCN, designed for time-evolving graph path classification, integrating temporal \\ dependencies and graph dynamics by relational GCN to process time-based relations.
\end{tabular}\\ \hline
DCRNN           & \cite{2017DCRNN}    & 2017 &  Flow  & road network & GCN & RNN &  \begin{tabular}[c]{@{}l@{}}
DCRNN models forecasting as a diffusion process on directed graphs, using bidirec-\\tional random walks and an encoder-decoder architecture with scheduled sampling.
\end{tabular}\\ \hline
CAGRU           & \cite{khodabandelou2021link} & 2021 &  Speed  & road network & GAT & GRU & \begin{tabular}[c]{@{}l@{}}
CAGRU predicts traffic speed and identifies patterns using a convolutional attention-\\based neural network based on traffic flow data without relying on historical speed data.
\end{tabular}
\\ \hline
DMSTGCN            & \cite{han2021dynamic} & 2021 &  Speed  & learned & DGNN & DGNN & 
\begin{tabular}[c]{@{}l@{}}
DMSTGCN learns dynamic spatial dependencies between road segments and incorpo-\\rates multi-varied traffic data, capturing multifaceted spatio-temporal traffic features.
\end{tabular}\\ \hline
FASTGNN         & \cite{zhang2021fastgnn} & 2021 &  Speed  & road network & ASTGCN & ASTGCN & 
\begin{tabular}[c]{@{}l@{}}
FASTGNN, a federated learning framework, features a differential privacy-based me-\\thod to protect topological information and an innovative aggregation approach.
\end{tabular}\\ \hline
-      & \cite{liu2020graphsage} & 2020 &  Speed  & road network & GraphSAGE & - & 
\begin{tabular}[c]{@{}l@{}}
This paper uses GraphSAGE to forecast spatially heterogeneous traffic speed and impu-\\tes missing data for segment networks with nonlinear spatial-temporal correlations.
\end{tabular}\\ \hline
ATT-LSTM        & \cite{wu2020combined}   & 2020 &  Speed  & road network & GAT & LSTM & 
\begin{tabular}[c]{@{}l@{}}
Attention-based LSTM (ATT-LSTM), a short-term level prediction model, predicts \\traffic speed and imputes missing traffic data with a data preprocessing module.
\end{tabular}\\ \hline
GATCN           & \cite{guo2020short}     & 2020 &  Speed  & road network & GAT & TCN & 
\begin{tabular}[c]{@{}l@{}}
GATCN, a deep learning framework combining GAT and TCN, effectively learns spatio-\\ temporal traffic flow characteristics and neighborhood information with multiple layers.
\end{tabular}\\ \hline
MTL-GRU         & \cite{zhang2020multitask} & 2020 &  Speed  & road network & GNN & GRU & \begin{tabular}[c]{@{}l@{}}
MTL-GRU, a multitask learning GRU model with residual mappings, selects \\ the most informative features to enhance traffic flow and speed forecasting.
\end{tabular}\\ \hline
DSTL-GR         & \cite{wang2023dynamic} & 2023 &  Time  & road network & GraphSAGE & LSTM & \begin{tabular}[c]{@{}l@{}}
DLSF-GR enhances travel time prediction by considering spatial and temporal dep-\\endence, as well as exogenous variables, through a combination of GNNs and RNNs.
\end{tabular}\\ \hline
DeepTrans       & \cite{tran2020deeptrans} & 2020 &   Time  & road network & DCRNN & DCRNN & \begin{tabular}[c]{@{}l@{}}
DeepTRANS enhances travel time estimation by incorporating traffic forecasting into \\ an existing deep learning-based bus ETA model, improving congestion prediction.
\end{tabular}\\ \hline
SST-GNN         & \cite{roy2021sst} & 2020 &  Time  & road network & SGNN & SGNN & \begin{tabular}[c]{@{}l@{}}
SST-GNN predicts by encoding spatial correlations, using neighborhood aggregati-\\on and a spatio-temporal mechanism with position encoding for periodic patterns.
\end{tabular}\\ \hline
-            & \cite{ma2019bus} & 2019 &  Time  & road segment & clustering & - & 
\begin{tabular}[c]{@{}l@{}}
The model predicts bus travel times using real-time taxi and bus data, dividing \\ routes into dwelling and transit segments with two tailored models for each.
\end{tabular}\\

\bottomrule
\end{tabular}
}
\label{tab:traffic_methods}
\vspace{-0.2in}
\end{table*}

\subsubsection{Traffic Flow Forecasting} \
Traffic flow forecasting, which uses flow as a metric, involves estimating the volume of vehicles during specific periods and across different segments of the transportation network. 
Accurate traffic flow forecasting is essential for managing congestion, planning routes, handling incidents, and improving overall transportation infrastructure efficiency. 
The challenges arise from the complex and constantly changing factors influencing traffic flow, such as daily commuting patterns, road conditions, weather, special events, accidents, and construction work \cite{zhao2023dynamic}. 

Recent research has introduced various models that can effectively capture the complex interdependencies in spatial and temporal data. 
These models, such as STSGCN  by Song et al. \cite{2020STSGCN},  DyHSL by Zhao et al. \cite{zhao2023dynamic}, DCRNN by Li et al.\cite{2017DCRNN}, ASTGCN by Guo et al. \cite{2019ASTGCN}, utilize advanced neural network architectures to handle dynamic relationships within traffic systems. 
They provide a comprehensive understanding of traffic behavior by addressing spatial aspects (e.g. road connectivity, road intersections) and temporal factors (e.g. traffic flow variations) over time.

One line of research involves utilizing GNNs in combination with RNNs \cite{hochreiter1997long, 2014GRU} to recursively capture both spatial and temporal information \cite{2017DCRNN, bai2020AGCRN, yi2020hypergraph}. 
For instance, the Diffusion Convolutional Recurrent Neural Network (DCRNN) by Li et al. \cite{2017DCRNN} replaces fully connected layers in the GRU \cite{2014GRU} with diffusion convolution. 
Similarly, the Adaptive Graph Convolutional Recurrent Network (AGCRN) by Bai et al. \cite{bai2020AGCRN} focuses on learning node-specific features and uncovering hidden inter-dependencies through an adaptive graph convolutional recurrent methodology. 
AGCRN reflects a trend toward tailoring models to understand complex network dynamics. 
Furthermore, the Hypergraph Convolutional Recurrent Neural Network (HGC-RNN) by Yi et al. \cite{yi2020hypergraph} combines hypergraph convolution with RNNs, specifically targeting traffic flow forecasting. This combination highlights the potential of integrating different neural network architectures to improve prediction accuracy.

Another line of research \cite{2020STSGCN, 2019ASTGCN, li2019predicting, zhao2023dynamic, ju2024cool} involves developing a large spatio-temporal graph and utilizing GNNs to capture spatio-temporal correlations. 
For example, the Spatial-temporal Synchronous Graph Convolutional Network (STSGCN) by Song et al. \cite{2020STSGCN} creates a spatio-temporal graph structure to perform localized graph convolution operations, thereby improving data processing capabilities. 
Attention-based Spatial-temporal Graph Convolutional Network (ASTGCN) by Guo et al. \cite{2019ASTGCN} incorporates an attention mechanism within the spatio-temporal graph context \cite{2020STSGCN} to enhance performance by focusing on salient features. 
Additionally, the Long Short-Term Memory R-GCN (LRGCN) approach by Li et al. \cite{li2019predicting} is designed to encode spatio-temporal graphs more efficiently, addressing the inherent complexities of such data structures.

The third line of research involves utilizing GNNs in conjunction with transformers \cite{xu2020spatial, huo2023hierarchical, zhang2023fptn, chen2022bidirectional}, leveraging the success of transformers across various domains \cite{vaswani2017attention,he2021transrefer3d, zhao2022target}. 
This approach effectively merges the transformers' ability to handle long-range dependencies with the spatial processing strengths of GNNs, resulting in two distinct types of methodologies for traffic flow forecasting. The first type features modular integration where spatial and temporal components are distinctly processed and then combined. For example, the Spatial-Temporal Transformer Networks (STTNs) introduced by Xu et al. \cite{xu2020spatial} employ a spatial transformer to dynamically model directed spatial dependencies via self-attention, capturing real-time traffic conditions and directional flow. This is complemented by a temporal transformer that handles long-range bidirectional temporal dependencies, optimizing long-term forecasting accuracy. Similarly, the Multi-Spatial-Temporal Encoder-Decoder Model (MST-EDM) \cite{zhang2023traffic} processes different temporal scales with distinct encoders that are later fused, offering a granular analysis of temporal correlations.

The second type adopts a more unified or pure transformer approach, leveraging a streamlined architecture to enhance processing efficiency and scalability. The Fast Pure Transformer Network (FPTN) \cite{zhang2023fptn}, for instance, restructures traffic flow data analysis by aligning it along the sensor dimension, applying multiple layers of Transformer encoders to simultaneously capture complex spatio-temporal correlations. This method significantly reduces computational time and resource usage. Additionally, the Bidirectional Spatial-Temporal Adaptive Transformer (Bi-STAT) \cite{chen2022bidirectional} uses an adaptive encoder-decoder architecture that dynamically adjusts to the complexity of traffic forecasting tasks. It integrates innovative features like a dynamic halting module (DHM) for iterative computation, enhancing the adaptability and accuracy of the forecasts.

The diversity of these models demonstrates the breadth of innovation in this area. Each approach provides unique insights and methodologies, contributing to an extensive and more diverse toolkit for traffic analysts and urban planners.

\subsubsection{Traffic Speed/Time Forecasting}
Traffic speed forecasting and time forecasting are closely intertwined within ITS, playing vital roles in improving navigational routing and estimating arrival times in various applications.
Traffic speed forecasting involves estimating the average velocity of vehicles over a specific segment during a defined time interval. In contrast, traffic time forecasting predicts the duration required to travel through a particular route or segment.
Due to the intrinsic relationship between travel speed and travel time, both forecasting types face similar challenges, such as information scarcity and heterogeneous traffic conditions. 
Recent research in this field \cite{han2021dynamic, liu2020graphsage, zhang2021fastgnn, mi2022dynamic, wu2020combined, guo2020short, khodabandelou2021link, zhang2020multitask, kang2020urban, tran2020deeptrans, abellana2023multivariate, wang2023dynamic, roy2021sst}, has effectively utilized both spatial and temporal dimensions in traffic speed data, similar in traffic flow forecasting.

\textbf{Information scarcity} is a biggest problem in these fields, which highlights the difficulty in generating accurate predictions when faced with limited, incomplete, or sparse traffic data \cite{huang2018sparse, wu2020combined}.
Such scarcity can stem from various reasons, including the lack of coverage by sensor networks, the high costs associated with the deployment and maintenance of extensive traffic monitoring systems, and the challenges in collecting data on roads with low traffic volumes or in remote areas. 
In response to these challenges, Liu et al. \cite{liu2020graphsage} have developed a technique that applies a data recovery algorithm based on identifying nonlinear spatial and temporal correlations within the road network. This algorithm helps impute missing speed data for different segments and enables traffic speed forecasting across a diverse and heterogeneous road network. 
Moreover, Huang et al. \cite{huang2018sparse} have utilized Probabilistic Principal Component Analysis (PPCA) to model travel speeds reliably, even when data is missing from specific road segments. 
They have also employed spectral clustering to categorize roads with similar traffic conditions into clusters, which reduces the variability of traffic conditions within each group. This enhances predictive consistency and facilitates parallel computing to improve overall prediction performance.

In traffic speed forecasting, Liu et al. \cite{liu2020graphsage} employs the GraphSAGE model \cite{hamilton2017inductive}, a novel approach tailored for sparse network conditions, to enhance the accuracy of traffic speed predictions. This approach emphasizes the importance of spatial information in the context of sparse connectivity.
Khodabandelou et al. \cite{khodabandelou2021link} proposed CAGRU, combining graph convolution techniques with attention-based gated recurrent units \cite{2014GRU} to capture both spatial and temporal relationships within traffic speed data. This fusion approach enriches the model's understanding of complex traffic dynamics. 
Zhang et al. \cite{zhang2020multitask} proposed a multi-task learning framework (MTL-GRU) that simultaneously processes traffic flow and speed forecasting. This approach enables the model to learn from the intertwined nature of traffic speed and flow, leading to a more nuanced representation of spatio-temporal data and enhancing the predictive accuracy for both metrics.

In traffic time forecasting, Tran et al. \cite{tran2020deeptrans} have taken the lead in this field by incorporating traffic flow forecasting models into their travel time prediction system, called DeepTrans. Their methodology uses machine learning to examine vast datasets of historical traffic patterns, allowing for more precise travel time estimations. 
Diving deeper into the interplay between spatial and temporal factors, Kang et al. \cite{kang2020urban} introduced a novel spatio-temporal forecasting framework focused on the urban context. 
This approach can process and integrate multifaceted data streams, capturing the intricate dynamics of urban traffic. The model considers not only the physical layout of the transportation network but also the fluctuating congestion levels over time. By assimilating this spatio-temporal information, their model extracts essential representations that significantly improve the reliability of travel time forecasts.

Although traffic speed/time forecasting and traffic flow forecasting are related, they are still distinct areas in transportation domains. 
Traffic flow forecasting offers a macroscopic view, focusing on overall traffic conditions and trends across a broader area or network \cite{zhao2023dynamic, 2017DCRNN, 2020GMAN}, which involves understanding traffic patterns, volume, and congestion across a network. 
On the other hand, traffic speed/time forecasting delves into the microscopic details, emphasizing the temporal elements of travel. It provides detailed insights into the travel duration between specific locations, making it valuable for journey planning and management \cite{wei2007development, barcelo2010travel, comi2020bus}.
More details about the related work for traffic forecasting can be found in Table.\ref{tab:traffic_methods}.

\subsubsection{Discussion}
\textbf{How to Construct a Graph in Traffic Forecasting?}
Contemporary methods primarily align with two categories: \textit{empirical graph construction} and \textit{learning-based graph construction}. Empirically constructed graphs, such as road networks, are commonly used \cite{bai2020AGCRN, 2017DCRNN, 2019ASTGCN, zhang2023fptn}. Techniques like dynamic time wrapping also play a role in capturing dependencies among time series \cite{li2021spatial, li2022weighted}.
Recent advancements have introduced methods to learn underlying graph structures from spatio-temporal data \cite{lan2022dstagnn, li2021spatial, zhao2023dynamic, jin2023transferable, zhang2020spatio}. Some approaches \cite{li2021spatial, jin2023transferable} focus on dissecting and understanding spatial and temporal structures individually, then integrating them for advanced spatio-temporal forecasting. For instance, Li et al. \cite{li2021spatial} proposed Spatial-Temporal Fusion Graph Neural Networks (STFGNN), which employ a spatial fusion graph along with a temporal graph, showcasing the efficacy of multi-dimensional graph structures in analysis. In contrast, other studies \cite{zhao2023dynamic, lan2022dstagnn, zhang2020spatio} aim to simultaneously model spatio-temporal structures. Zhao et al. \cite{zhao2023dynamic} introduced a dynamic hypergraph over the spatio-temporal graph to enhance forecasting accuracy. Meanwhile, the Dynamic Spatial-Temporal Aware Graph Neural Network (DSTAGNN) \cite{lan2022dstagnn} focuses on learning integrated spatio-temporal graphs, utilizing multi-head attention \cite{vaswani2017attention} to capture dynamic spatial relevancies.

\textbf{How standard GNNs are modified for traffic forecasting?}
Traffic forecasting involves predicting future traffic patterns by leveraging temporal information for accuracy. 
This process is divided into two sub-types, both of which utilize similar methodologies and share the fundamental principle that integrating spatio-temporal information is essential for enhancing model performance.
To achieve this, a combination of GNNs \cite{hamilton2017inductive, kipf2016semi, xu2019powerful} and sequential models such as RNNs, Long Short-Term Memory networks (LSTMs) \cite{hochreiter1997long, FC-LSTM}, and Gated Recurrent Units (GRUs) \cite{2014GRU} is commonly employed. 
Sequential models excel at processing time-series data, making them particularly effective for capturing and forecasting time-dependent traffic patterns. 
When combined with GNNs, which model spatial relationships, this approach adeptly captures both temporal sequences and spatial dependencies within traffic data, resulting in a more accurate and comprehensive traffic forecasting model.

\subsection{Vehicle Control System}

\begin{table*}[]
\centering
\caption{A Comprehensive Overview of Most Related Studies for Vehicle Control System}
\label{table:vehicle_control}
\resizebox{\textwidth}{!}{%
\begin{tabular}{ccclcl}
\hline
\multicolumn{1}{c}{\textbf{Model}} &
\multicolumn{1}{c}{\textbf{Article}} &
\multicolumn{1}{c}{\textbf{Year}} &
\multicolumn{1}{c}{\textbf{Datasets}} &
\multicolumn{1}{c}{\textbf{GNN Module}} &
\multicolumn{1}{c}{\textbf{Summary}} \\ 
\hline
GTNet  
&\cite{zhou2023gtnet} 
&2023
&\begin{tabular}[c]{@{}l@{}}ModelNet40, ShapeNet part\end{tabular} 
&Graph Transformer
&\begin{tabular}[c]{@{}l@{}} 
GTNet uses a Local Transformer to calculate neighboring point weights through dynamic graph-based \\cross-attention within domains, and a Global Transformer to expand its range using global self-attention.
\end{tabular}  
\\
\hline

MHNet
&\cite{liu2023multi}
&2023
&\begin{tabular}[c]{@{}l@{}}ModelNet40, NTU\end{tabular} 
&Spectral GNN  
&\begin{tabular}[c]{@{}l@{}} 
MHNet introduces a polynomial hypergraph filter, which dynamically extracts multi-scale node features.
\end{tabular}  
\\
\hline

DiffConv
&\cite{lin2022diffconv} 
&2022
&\begin{tabular}[c]{@{}l@{}}ModelNet40, Toronto3D,\\ ShepeNet part\end{tabular} 
&Spectral GNN  
&\begin{tabular}[c]{@{}l@{}} 

DiffConv uses density-dilated neighborhoods where each point's radius depends on its kernel density. \\It also uses masked attention to introduce task-specific learned variations to the neighborhood.
\end{tabular}  
\\
\hline

DeltaConv
&\cite{wiersma2022deltaconv}
&2022
&\begin{tabular}[c]{@{}l@{}}ModelNet40, SHREC11,\\  ScanObjectNN, ShapeNet\end{tabular} 
&Spectral GNN  
&\begin{tabular}[c]{@{}l@{}} 
DeltaConv uses a graph-based anisotropic convolutional operator by combining a set of geometric \\ operators defined on scalar and vector fields to encode the directional information of each surface point.
\end{tabular}  
\\
\hline

3DCTN 
&\cite{lu20223dctn}
&2022
&\begin{tabular}[c]{@{}l@{}}ModelNet40, ScanObjectNN\end{tabular} 
&Graph Transformer
&\begin{tabular}[c]{@{}l@{}} 
3DCTN combines convolutions and transformers to learn local and global features. It uses a multi-scale \\ local feature aggregation block and a global feature learning block to process downsampled point sets.
\end{tabular}  
\\
\hline

Point Transformer
&\cite{zhao2021point}
&2021
&\begin{tabular}[c]{@{}l@{}}S3DIS, ModelNet40,\\ShapeNet part\end{tabular} 
&Graph Transformer
&\begin{tabular}[c]{@{}l@{}} 
Point Transformer introduces an expressive transformer layer tailored for point cloud processing. \\ It employs local self-attention and integrates vector attention to achieve elevated accuracy levels.
\end{tabular}  
\\
\hline

PCT
&\cite{guo2021pct}
&2021
&\begin{tabular}[c]{@{}l@{}}ModelNet40, ShapeNet\end{tabular} 
&Graph Transformer
&\begin{tabular}[c]{@{}l@{}} 
Point Cloud Transformer (PCT) improves capturing local context capture within the point cloud by using \\ coordinate-based input embedding with the help of farthest point sampling and nearest neighbor search. 
\end{tabular}  
\\
\hline

CurveNet
&\cite{xiang2021walk}
&2021
&\begin{tabular}[c]{@{}l@{}}ModelNet40, ModelNet10, \\ ShapeNet part\end{tabular} 
&Spatial GNN  
&\begin{tabular}[c]{@{}l@{}} 
CurveNet enhances point cloud shape descriptors by organizing connected points through guided \\ walks within point clouds and aggregating them to enhance their individual point-wise features.
\end{tabular}  
\\
\hline

LDGCNN
&\cite{zhang2021linked} 
&2021
&\begin{tabular}[c]{@{}l@{}}ModelNet40, ShapeNet\end{tabular} 
&Spatial GNN  
&\begin{tabular}[c]{@{}l@{}} 
LDGCNN is a linked dynamic graph CNN created for direct classification and segmentation of point clouds, \\ addressing sparsity and unstructured nature. It also includes theoretical analysis and model visualization.
\end{tabular}  
\\
\hline

3D-GCN
&\cite{lin2020convolution} 
&2020
&\begin{tabular}[c]{@{}l@{}}ModelNet40, ModelNet10, \\ ShapeNet part\end{tabular} 
&Spatial GNN  
&\begin{tabular}[c]{@{}l@{}} 
3D-GCN is a novel approach for processing 3D point clouds in computer vision that offers scale and shift \\ invariance by utilizing learnable kernels and a graph max-pooling mechanism to extract robust features.
\end{tabular}  
\\
\hline

DHGNN
&\cite{feng2019hypergraph}
&2019
&\begin{tabular}[c]{@{}l@{}}ModelNet40, NTU\end{tabular} 
&Spectral GNN  
&\begin{tabular}[c]{@{}l@{}} 
DHGNN addresses limitations in graph/hypergraph-based deep learning by dynamically updating hyper-\\graph structures and encoding high-order data relations through vertex and hyperedge convolutions.
\end{tabular}  
\\
\hline

DGCNN
&\cite{wang2019dynamic} 
&2019
&\begin{tabular}[c]{@{}l@{}}ModelNet40\end{tabular} 
&Spatial GNN  
&\begin{tabular}[c]{@{}l@{}} 
DGCNN, a novel neural network module dubbed EdgeConv suitable for point clouds, enhances CNN-\\based high-level tasks by incorporating local neighborhood information and adapting to topology.
\end{tabular}  
\\
\hline

RGCNN
&\cite{te2018rgcnn}
&2018
&\begin{tabular}[c]{@{}l@{}}ShapeNet part\end{tabular} 
&Spectral GNN  
&\begin{tabular}[c]{@{}l@{}} 
RGCNN directly processes point clouds, utilizing spectral graph theory and Chebyshev polynomial \\ approximation to capture dynamic graph structures adaptively, enhancing point cloud understanding.
\end{tabular}  
\\
\hline

AGCN
&\cite{li2018adaptive}
&2018
&\begin{tabular}[c]{@{}l@{}}Sydney urban\end{tabular} 
&Spectral GNN  
&\begin{tabular}[c]{@{}l@{}} 
AGCN, a flexible Graph CNN that takes data of arbitrary graph structure as input, enables task-driven \\ adaptive graph and distance metric learning for diverse data such as molecular and social networks.
\end{tabular}  
\\
\hline

KCNet
&\cite{shen2018mining} 
&2018
&\begin{tabular}[c]{@{}l@{}}ModelNet40, ShapeNet\end{tabular} 
&Spatial GNN  
&\begin{tabular}[c]{@{}l@{}} 
KCNet improves semantic learning efficiency for 3D point clouds by introducing a point-set kernel for 3D \\ geometry and recursive feature aggregation on a nearest-neighbor graph that focuses on local structures.
\end{tabular}  
\\
\hline

Local-SpecGCN
&\cite{wang2018local}
&2018
&\begin{tabular}[c]{@{}l@{}}ModelNet40, McGill Shape,\\ShapeNet part, \\ ScanNet Indoor Scene\end{tabular} 
&Spectral GNN  
&\begin{tabular}[c]{@{}l@{}} 
Local-SpecGCN uses spectral graph convolution on local graphs and a graph pooling strategy for point\\  cloud feature learning, enhancing feature descriptors by aggregating information from clustered nodes.
\end{tabular}  
\\
\hline

ECC 
&\cite{simonovsky2017dynamic} 
&2017 
&\begin{tabular}[c]{@{}l@{}}Sydney Urban Objects, \\ModelNet10, ModelNet40\end{tabular} 
&Spatial GNN  
&\begin{tabular}[c]{@{}l@{}}  
ECC adapts convolution operators for arbitrary graphs, avoiding the spectral domain, and uses specific edge \\ labels in a vertex's neighborhood to condition filter weights, enabling diverse graph classification tasks.
\end{tabular}
\\
\hline
\end{tabular}%
}
\vspace{-0.15in}
\end{table*}

Vehicle Control System (VCS) is an essential component of ITS, which is responsible for optimizing the operation, safety, and efficiency of autonomous vehicles in traffic networks. 
Recently, learning-based methods, especially GNNs, have demonstrated significant potential in addressing the complexities of urban environments. This section will delve into two key subsystems within VCS: perception, which involves interpreting sensory data, and trajectory prediction, which focuses on forecasting vehicle movement.
Finally, we present an overview of the most related studies for VCS in Table.\ref{table:vehicle_control}.

\subsubsection{Perception}

Perception plays a vital role in vehicle control system in identifying and categorizing objects around a vehicle. 
Perception involves two critical tasks: semantic segmentation with classification and object detection with tracking \cite{jebamikyous2022autonomous}, which includes clustering and assigning specific classes to pixels in an image.
In perception, point cloud is a common method for representing 3D data. It can capture complex 3D shapes and their unique irregular structures.
Traditional deep learning methods usually convert point clouds into 3D voxel grids or collections of images before feeding them into deep neural networks, potentially resulting in information loss and computational overhead \cite{guo2020deep}. 
An alternative approach leverages the graph-like nature of point clouds, leading to a growth in research efforts employing GNNs to enhance the efficiency and accuracy of 3D data analysis. 
In the following sections, we will discuss GNN-based methods for learning representations from point cloud data.

\textbf{Graph-Based Methods in Spatial Domain.} 
Spatial Convolutional Graph Neural Networks can be described as the process of spreading node features to neighboring nodes by utilizing a convolutional kernel. This is then followed by applying an activation function using a trainable weight matrix to transform these features into the subsequent hidden layer ~\cite{balcilar2020bridging}. 

As a pioneering approach, Simonovsky et al. \cite{simonovsky2017dynamic} introduced Edge-Conditioned Convolution (ECC) as the first graph-based method in a spatial domain. This method uses edge labels in vertex neighborhoods to calculate adaptive convolution kernel weights, thus enabling more effective utilization of edge information compared to traditional point-based convolutions.
However, ECC \cite{simonovsky2017dynamic} has a limitation in that it primarily depends on the inherent graph structure of the input point cloud, which inherently restricts its flexibility to model non-local relations. 
To address the limitation, several methods ~\cite{wang2019dynamic,zhang2021linked, wang2021object} have been proposed.

Dynamic Graph Convolutional Neural Network (DGCNN) \cite{wang2019dynamic} introduces an EdgeConv neural network architecture to segment point clouds and capture semantically related structures. 
This approach learns a dynamic graph representation of the point cloud, which evolves across layers and during the training phase as learnable parameters are updated.
Building upon earlier developments like ECC and DGCNN, the Linked Dynamic Graph Convolutional Neural Network (LDGCNN) \cite{zhang2021linked} enhances the capabilities of DGCNN by establishing links between hierarchical features derived from various dynamic graphs. This linkage enables the computation of informative edge vectors while simultaneously reducing the model's size.

To capture the local neighborhood structural information of a point, kernel-based approaches have been extensively explored ~\cite{shen2018mining, lin2020convolution, xiang2021walk}. 
For instance, KCNet \cite{shen2018mining} introduced a point-set kernel of learnable 3D points. They utilized a kernel correlation layer to compute affinities between each data point's nearest neighbors and these point-set kernels. 
They also implemented recursive feature propagation and aggregation along the edges, effectively utilizing local high-dimensional feature structures.
Similarly, 3D-GCN \cite{lin2020convolution} proposed deformable kernels that were designed to extract shift and scale-invariant local 3D features from point clouds. 
Furthermore, Xiang et al. \cite{xiang2021walk} introduced a method to arrange connected points through guided walks within the point clouds and subsequently aggregated them to enhance their point-wise features, effectively improving the representation of point cloud geometry.

\textbf{Graph-Based Methods in Spectral Domain.} Spectral Convolutional Graph Neural Networks are based on spectral graph theory \cite{chung1997spectral}. In this framework, graph signals are filtered through the eigendecomposition of the graph Laplacian. 
Regularized Graph CNN (RGCNN) \cite{te2018rgcnn} perform graph convolution and feature learning based on spectral graph theory, which treats point cloud features as signals on a graph and employs Chebyshev polynomial approximation for graph convolution. 
RGCNN adapts to the corresponding learned features by updating the graph Laplacian matrix in each layer, effectively capturing evolving graph structures during the learning process. 
Traditional spectral GCNs require the prior computation of graph Laplacians and pooling hierarchies for the entire graph, which can be computationally intensive.
In dealing with the challenges of diverse graph topology in data, two promising approaches have been proposed. 
One approach is the Adaptive Graph Convolutional Neural Network (AGCN) \cite{li2018adaptive}, which enhances the generalization capacity of GCNs by incorporating a learnable distance metric to parameterize the similarity between two vertices within a graph, allowing for the dynamic construction of graphs. 
The other approach is Local-SpecGCN \cite{wang2018local}, which conducts spectral filtering on dynamically generated local graphs. It uses recursive clustering based on spectral coordinates to facilitate graph pooling, which enhances the learning process by mitigating point isolation. 
Instead of using conventional max pooling, Wang et al. introduced a recursive clustering and pooling strategy that enables the amalgamation of information from nodes within clusters defined by their spectral coordinates.

Hypergraphs attract the attention of researchers as a tool for capturing high-order data correlations. 
One notable example is Hypergraph Neural Networks (HGNN) \cite{feng2019hypergraph}, which uses a hyperedge convolution operation to capture high-order data correlations and represent complex structures within point clouds. This operation aggregates node features into hyperedge features and then updates node features through hyperedge feature aggregation. 
Hypergraph Gragh Convolutional Network (HyperGCN) \cite{yadati2019hypergcn} uses non-linear Laplacian operators \cite{chan2020generalizing} to convert hypergraphs into more straightforward graphs by breaking hyperedges down into subgraphs with edge weights that depend solely on their degrees. 
Hypergraph convolution relies on a predefined structure for propagation. To address this limitation, Bai et al. \cite{bai2021hypergraph} introduced an attention mechanism for dynamic connection learning among hyperedges. This mechanism ensures that information propagates and gathers in graph regions relevant to specific tasks, leading to the learning of more discriminative node embeddings. 
Multi-modal Hypergraph Neural Network (MHNet) \cite{liu2023multi} uses hypergraph structures to model high-order and multi-modal data correlations effectively. It achieves this by employing a polynomial hypergraph filter that dynamically extracts multi-scale node features through parametric polynomial fitting.

Recent advancements have been made in convolution operations for point clouds.
Conventional approaches commonly perform convolution operations to the irregular point clouds by imposing a fixed view, such as using fixed neighborhood sizes. 
To address this issue, DiffConv \cite{lin2022diffconv} introduced density-dilated neighborhoods, where the radius for each point depends on its kernel density.
DiffConv uses masked attention, which introduces task-specific irregularity to the neighborhood, making the convolution more flexible and effective. 
Another approach, DeltaConv. \cite{wiersma2022deltaconv} proposed a new method to construct anisotropic convolution layers for geometric CNNs.
It designed a graph-based anisotropic convolutional operator by combining a set of geometric operators defined on scalar and vector fields to encode directional information for each surface point.

\textbf{Graph Transformer-based Methods}.
While transformers have been extensively used in computer vision, graph-based transformers are explicitly tailored for learning 3D point cloud representation. 
The transformer architecture is well-suited for point cloud analysis due to its self-attention operator, which functions as a set operator by preserving permutation and cardinality invariance of input elements \cite{zhao2021point}. 
As an example within this category, Point Transformer (PT)\cite{zhao2021point} introduces a transformer layer that is highly expressive and specifically designed for point cloud processing. The Point Transformer employs local self-attention that ensures scalability even in large scenes. Additionally, integrating vector attention is pivotal in achieving elevated accuracy levels. 
Another tailored transformer for point clouds is  Point cloud transformer (PCT) \cite{guo2021pct}. PCT employs a coordinate-based input embedding module to learn distinctive features by combining raw positional encoding and input embedding, harnessing the individual spatial coordinates of each point. Furthermore, it enhances performance by substituting the original self-attention module with an offset-attention module. Unlike PT, PCT excels in capturing global interaction and local neighborhood information.

To improve efficiency in point cloud classification, 3D Convolution-Transformer Network (3DCTN) \cite{lu20223dctn} by Lu et al. combines GNN convolutions with transformers, which helps effectively learn local and global features. 
To be more specific, 3DCTN processes downsampled point sets using a combination of a multi-scale local feature aggregating block and a global feature learning block, which are implemented by GNNs and Transformers, respectively. 
While most Transformer-based approaches primarily rely on global attention mechanisms to extract point cloud features, they often fail to adequately capture feature learning from local neighbors. 
To address this challenge, Graph Transformer Network (GTNet) \cite{zhou2023gtnet} employs local and global Transformer modules. 
The local Transformer module calculates neighboring point weights through dynamic graph-based intra-domain cross-attention, which assigns different weights to each neighboring point's influence on the centroid's features. 
In contrast, the global Transformer module expands the local Transformer's reach by utilizing global self-attention, which enables a broader feature extraction.

\subsubsection{Trajectory Prediction} 
Trajectory prediction involves anticipating the future paths of road users based on their past trajectories and the surrounding environment. Road users include vehicles, cyclists, and pedestrians. The environment includes static factors such as terrain and obstacles and dynamic factors such as the movements of nearby agents \cite{huang2023multimodal}.

Several models have been developed to improve trajectory prediction by adopting the paradigm of spatial and temporal convolution through GNNs. 
For instance, GRIP \cite{li2019grip} enhanced trajectory prediction by incorporating interactions among adjacent objects, represented as an undirected graph. It employs a GCN module to model the graph network, and the output of GCN is then input into an LSTM encoder-decoder for predicting the trajectories of surrounding vehicles. 
SCALE-Net \cite{jeon2020scale} creates an efficient and scalable framework to maintain high prediction performance for numerous vehicles. It employs an Edge-Enhanced Graph Convolutional Network (EGCN) to update node features based on an attention mechanism influenced by edge features from neighboring nodes. 
Social-STGCNN \cite{mohamed2020social} represents pedestrian trajectories as spatio-temporal graphs and employs GCN and TCN to operate on these graphs, enabling the model to predict the entire sequence simultaneously. 
Chandra et al. \cite{chandra2020forecasting} uses a two-layer Graph-LSTM architecture for trajectory prediction. 
The initial layer is applied to forecast future trajectories of traffic participants, while the second layer captures interaction-related factors among participants using a weighted dynamic geometric graph network (DGG). 
Additionally, they introduced a regularization algorithm based on spectral clustering to minimize errors in long-term predictions.
GSTCN \cite{sheng2022graph} uses a GCN to capture spatial interactions and a CNN to handle temporal correlations among neighboring vehicles. The spatial-temporal features are encoded and decoded using a GRU network in their framework.

Recently, the attention mechanism has gained popularity for various sequence-based tasks, including predicting the trajectory of autonomous vehicle systems.
The Spatial-Temporal Graph Attention network (STGAT) \cite{huang2019stgat} utilizes an LSTM encoder to encode trajectories, followed by GAT for attention-weighted interaction information, and an LSTM decoder for trajectory prediction.
SCOUT \cite{carrasco2021scout} utilizes GAT to incorporate dynamic agent interactions, aiming to improve socially aware and consistent trajectory predictions.
The Attention-based Spatio-Temporal Graph Neural Network (AST-GNN) \cite{zhou2021ast} utilizes a dual-attention mechanism: the first for capturing spatial interactions among all agents and the second for considering the temporal movement patterns of each agent in the past. 
The Spatio-Temporal Graph Dual-Attention Network (STG-DAT) \cite{li2022spatio} employs a dual-attention mechanism to learn representations on spatio-temporal dynamic graphs, considering historical and future features from state, relation, and scene context information.
The Triple Policies Fused Hierarchical Graph Networks (Tri-HGNN) \cite{zhu2023tri} proposed triple policies fused hierarchical GNN for pedestrian trajectory prediction. More specifically, the extrinsic-level policy uses GAT for spatial and temporal embeddings, the intrinsic-level policy captures human intention with GCN, and the basic-level policy combines information for predictions through TCN.
The Heterogeneous Driving Graph Transformer (HDGT) \cite{jia2023hdgt} models the driving scene as a heterogeneous graph, where agents, lanes, and traffic signs are considered as different types of nodes and edges. Besides, the transformer structure is applied hierarchically to accommodate the heterogeneous inputs.

\subsubsection{Discussion}

\textbf{How to Construct a graph in Vehicle Control System?}
In vehicle perception, 3D representations of real-world objects can be obtained using 3D sensors or Light Detection and Ranging (LiDAR) technology. The individual points in a point cloud can be seen as nodes in a graph, while determining the graph structure is a complex process.
Connecting all edges for the entire point cloud can require a lot of memory. Therefore, a simplified method involves using the K-Nearest Neighbors (KNN) approach to create a locally directed graph \cite{zhang2021linked}. Moreover, there have been proposals for using kernel-based methods to calculate the affinities between each data point ~\cite{shen2018mining, lin2020convolution, xiang2021walk}.
In trajectory prediction, both the agent (such as a vehicle or pedestrian) and static features in the environment (such as traffic signals and road signs) can be viewed as nodes in a graph. 
The connections between these nodes can represent their physical proximity or potential interaction. Node attributes may encompass the entity's speed, acceleration, heading, and previous path, while edge properties may include relative speed, relative heading, and distance between the two entities. To account for temporal aspects, temporal encoding methods can be utilized, such as using time steps as node or edge attributes or employing recursive GNN variations for processing time series data.

\textbf{How standard GNNs are modified for Vehicle Control System?}
When GNNs are deployed in real-time and high-stakes environments such as VCS, adapting standard GNN architectures is essential to meet specific operational demands. 
One key challenge is the dynamic nature of the traffic environment, where the graph structure frequently changes. To address this, methods like edge weighting are crucial for dynamically updating the graph structure without retraining the model from scratch.
Another important consideration is incorporating both spatial and temporal awareness. 
Spatial-temporal GNNs are often employed to integrate spatial relationships and temporal dynamics into the model, which enhances its ability to understand and predict traffic patterns.
Additionally, autonomous vehicles require real-time or near-real-time responses, making the computational efficiency of GNNs critical. Standard GNNs can be computationally intensive, especially with large graphs, so optimizing the network for low latency is essential. This can be achieved through model simplification, pruning, and reducing the number of layers to accelerate computation.
Finally, in high-stakes scenarios, GNNs must account for uncertainty in their predictions. Techniques such as Bayesian GNNs \cite{zhou2020variational} can provide a measure of confidence, which is vital for safe and reliable decision-making.

\subsection{Traffic Signal Control}
Traffic signal control (TSC) involves managing and coordinating traffic lights at intersections and road junctions to regulate traffic flow, improve road safety, and minimize delays. 
Effective TSC helps reduce the likelihood of accidents by managing the movement of vehicles and pedestrians. It also has the potential to reduce travel time and fuel consumption, leading to lower vehicle emissions and supporting environmental sustainability. 
However, the intricate interactions between intersections and the continuously changing traffic conditions make the real-world network of intersections highly complex, posing a significant challenge for adaptive traffic signal control.

\begin{table*}[]
\centering
\caption{A Comprehensive Overview of Most Related Studies for Traffic Signal Control}
\label{table:traffic_signal_control}
\resizebox{\textwidth}{!}{%
\begin{tabular}{ccclccccl}
\hline
  \multicolumn{1}{c}{\textbf{Model}} &
  \multicolumn{1}{c}{\textbf{Article}} &
  \multicolumn{1}{c}{\textbf{Year}} &
  \multicolumn{1}{c}{\textbf{Datasets}} &
  \multicolumn{1}{c}{\textbf{Simulator}} &
  \multicolumn{1}{c}{\textbf{Temporal Module}} &
  \multicolumn{1}{c}{\textbf{Spatial Module}} &
  \multicolumn{1}{c}{\textbf{Attention Based}} &
  \multicolumn{1}{c}{\textbf{Summary}} \\ \hline
AFMRL &
  \cite{ma2022feudal} &
  2023 &
  \begin{tabular}[c]{@{}l@{}}Simulated and real-world data\\ (Jinan, Hangzhou, Manhattan)\end{tabular} &
  CityFlow &
  - &
  GNN &
  \XSolidBold &
  \begin{tabular}[c]{@{}l@{}} A multi-agent reinforcement learning approach for multi-intersection TSC. \\
  Adaptive partitioning is emphasized and feudal hierarchy is explored. \end{tabular} \\ \hline
KeyLight &
  \cite{lin2023keylight} &
  2023 &
  \begin{tabular}[c]{@{}l@{}}Simulated and real-world data\\ (Jinan, Hangzhou, New York)\end{tabular} &
  CityFlow &
  - &
  GAT &
  \CheckmarkBold &
  \begin{tabular}[c]{@{}l@{}} KeyLight integrates reinforcement learning and GNNs. NOV-LADLE \\ state representation and residual connections are used in the model. \end{tabular} \\ \hline
HG-M2I &
  \cite{yang2023hierarchical} &
  2023 &
  real world data (Chengdu) &
  SUMO &
  GRU &
  Bi-GRU &
  \CheckmarkBold &
  \begin{tabular}[c]{@{}l@{}} The HG-M2I algorithm, spatial-temporal analysis and multi-agent RL based, \\ optimizes TSC by hierarchical graph structures and input-output correlation.\end{tabular} \\ \hline
MetaSTGAT &
  \cite{wang2022meta} &
  2022 &
  \begin{tabular}[c]{@{}l@{}}Simulated and real-world data\\ (Jinan, Hangzhou)\end{tabular} &
  CityFlow &
  LSTM &
  GAT &
  \CheckmarkBold &
  \begin{tabular}[c]{@{}l@{}}MetaSTGAT, meta-learning based, merges GAT and LSTM to address \\ spatial-temporal correlations and dynamic interaction of intersections. \end{tabular} \\ \hline
PRGLight &
  \cite{zhao4040526dynamic} &
  2022 &
  \begin{tabular}[c]{@{}l@{}}Simulated and real-world data\\ (Jinan, Hangzhou, New York)\end{tabular} &
  CityFlow &
  - &
  GNN &
  \XSolidBold &
  \begin{tabular}[c]{@{}l@{}} PRCOL uses lane capacity for the RL reward function and GNN modules \\ to help RL decide the light phase and duration by predicting traffic flow.\end{tabular} \\ \hline
DynSTGAT &
  \cite{wu2021dynstgat} &
  2021 &
  \begin{tabular}[c]{@{}l@{}}Simulated and real world data\\ (Jinan, Hangzhou, New York)\end{tabular} &
  CityFlow &
  \begin{tabular}[c]{@{}c@{}}TCN, LSTM,\\  STGAT\end{tabular} &
  STGAT &
  \CheckmarkBold &
  \begin{tabular}[c]{@{}l@{}} DynSTGAT combines spatial-temporal graph attention networks \\ and temporal convolutional network to enhance adaptive TSC.\end{tabular} \\ \hline
IHG-MA &
  \cite{yang2021ihg} &
  2021 &
  \begin{tabular}[c]{@{}l@{}}Simulated and real-world data \\ (Chengdu)\end{tabular} &
  SUMO &
  Bi-GRU &
  Bi-GRU &
  \CheckmarkBold &
  \begin{tabular}[c]{@{}l@{}} IHG-MA uses inductive heterogeneous GNNs to capture traffic features \\ and a decentralized multi-agent actor-critic framework to optimize TSC.\end{tabular} \\ \hline
GraphLight &
  \cite{zeng2021graphlight} &
  2021 &
  Simulated data &
  SUMO &
  - &
  GCNN &
  \XSolidBold &
  \begin{tabular}[c]{@{}l@{}}GraphLight is a decentralized, graph-based, multi-agent system using actor-\\critic methods for TSC, distinguishing neighboring intersection impacts.\end{tabular} \\ \hline
TSC-GNN &
  \cite{zhong2021probabilistic} &
  2021 &
  \begin{tabular}[c]{@{}l@{}} real world data \\
  (Jinan, Hangzhou) \end{tabular}&
  - &
  - &
  GAT &
  \CheckmarkBold &
  \begin{tabular}[c]{@{}l@{}} TSC-GNN is a graph-based model for TSC utilizing probabilistic \\ neural networks, to manage uncertainties and calculate Q-values.\end{tabular} \\ \hline
STMARL &
  \cite{wang2020stmarl} &
  2020 &
  \begin{tabular}[c]{@{}l@{}}Simulated and real-world data\\ (Hefei, Hangzhou)\end{tabular} &
  CityFlow &
  RNN &
  GNN &
  \CheckmarkBold &
  \begin{tabular}[c]{@{}l@{}} STMARL applies spatial-temporal RL for TSC, using graphs, RNNs, \\GNNs, and deep Q-learning for distributed decision-making.\end{tabular} \\ \hline
CoLight &
  \cite{wei2019colight} &
  2019 &
  \begin{tabular}[c]{@{}l@{}}Simulated and real-world data\\ (Jinan, Hangzhou, New York)\end{tabular} &
  CityFlow &
  - &
  GAT &
  \CheckmarkBold &
  \begin{tabular}[c]{@{}l@{}}CoLight uses graph attention networks for TSC, and captures spatial-\\temporal impacts from nearby intersections without indexing\end{tabular} \\ \hline
\end{tabular}%
}
\vspace{-0.2in}
\end{table*}

\subsubsection{Multi-agent Graph Reinforcement Learning}
Single-agent reinforcement learning methods are constrained to managing traffic signals in a single intersection due to the curse of dimensionality when using a global single model for all intersections. 
As a result, single-agent RL is typically restricted to an isolated intersection without coordination with neighboring intersections ~\cite{shashi2021study}. 
In managing multiple intersections, a practical approach is to obtain neighborhood intersection information by combining the states of intersections and their adjacent areas  ~\cite{huo2020cooperative}. 
However, this method encounters difficulties when dealing with increasing input dimensionality, leading to challenges in model convergence.
Thankfully, multi-agent reinforcement learning allows for the individual control of each signal using a reinforcement learning (RL) agent and developing policies for each intersection, resulting in significant advancements. Additionally, the application of multi-agent graph reinforcement learning has demonstrated promising progress ~\cite{chu2019multi, gupta2017cooperative, zeng2021graphlight, zhao4040526dynamic, wu2021dynstgat}.


Nishi et al. \cite{nishi2018traffic} are among the ones who first combine multi-agent reinforcement learning and graph neural networks to address the multi-intersection interaction problem and the spatial dependency. Their work employs GCNs to extract the geometric features.
Zhong et al. \cite{zhong2021probabilistic} proposed a model named TSC-GNN to handle a problem that most studies model traffic state deterministically and to exploit the uncertainties of traffic conditions.
Yoon et al. \cite{yoon2021transferable} claimed that the RL method encountered a restricted exploration problem, which means it cannot handle unseen conditions. They proposed a novel approach to obtain a transferable policy by using graph representation for the state and training it by GNNs.
Based on Multi-Agent Reinforcement Learning (MARL), Saki et al. ~\cite{saiki2023flexible} used multi-objective reinforcement learning (MORL) to further improve the performance by determining the policy corresponding to each traffic flow ratio, which achieved the shorted average travel times in all environments compared with ruled based and single objective reinforcement learning.
Some more similar literature ~\cite{yang2021ihg, zeng2021graphlight, lin2023keylight, zhao4040526dynamic, wu2021dynstgat, yang2023hierarchical, ma2023learning} based on graph reinforcement learning is listed in Table.\ref{table:traffic_signal_control} .

\subsubsection{Attention Mechanism}
Another issue is distinguishing the impact of neighboring traffic signals on the target intersection. For example, intersections on the main road may significantly affect the target intersection more than those on the side road. Most existing research does not differentiate the influence of surrounding intersections on the target intersection~\cite{yu2021macar, nishi2018traffic}.
The attention mechanism has proven valuable in adaptive signal light control. 
CoLight \cite{wei2019colight} advanced this by using GATs to identify the impact of neighboring intersections and effectively leverage joint intersections. This was achieved by developing an index-free model of neighboring intersections and averaging their influences using learned attention parameters.
Sun et al. ~\cite{lin2023keylight}  introduced NOV-LADLE to address potential convergence failures in the attention mechanism, maintaining a concise state and emphasizing essential intersections. 
They also added a residual connection structure to GAT to accelerate convergence and enhance performance based on CoLight.
Further studies, including DynSTGAT and TSC-GNN \cite{wu2021dynstgat, zhong2021probabilistic}, have also explored using the graph attention mechanism to tackle this issue. Table \ref{table:traffic_signal_control} summarizes these works.

\subsubsection{Spatial and Temporal Dependency}
Similar to the previous discussion, considering spatial and temporal relations is crucial. The historical states of surrounding intersections affect the prediction of future signals at a target intersection.

Wang et al. \cite{wang2020stmarl} pioneered the study of spatio-temporal dependencies among multiple traffic signals. They used graph structures to capture spatial features and recurrent neural networks to integrate historical traffic data, with decisions for each signal based on deep Q-learning.
Li et al. ~\cite{li2020deep} proposed a model using LSTM and GCN to extract spatial-temporal traffic features of intersection networks. LSTM processed variable-length inputs and extracted valid features from historical data, while GCN handled the LSTM output, linking interactions of intersections. They incorporated imitation learning instead of reinforcement learning.
Yang~\cite{yang2023hierarchical} introduced the Hierarchical Graph Multi-agent Mutual Information (HG-M2I) algorithm to generate optimal embeddings of traffic networks. This algorithm fused multi-granularity information from each agent's current and historical states to develop optimal TSC policies, measuring the correlation between input states and output embeddings by maximizing mutual information.

Despite numerous studies attempting to incorporate temporal and spatial influences of surrounding intersections into the target intersection, they typically consider the spatial-temporal information separately. Wu et al. \cite{wu2021dynstgat} proposed DynSTGAT, which uses TCN to simultaneously capture historical and current spatial-temporal information.
To address dynamically changing traffic, Wang et al. \cite{wang2022meta} proposed MetaSTGAT, a meta-learning model based on GATs that adapts to dynamic traffic flow and fully utilizes the spatial-temporal characteristics of multi-intersections.
Other literature also explores the exploitation of spatial and temporal information~\cite{yang2021ihg, wang2022meta}.

\subsubsection{Discussion}
\textbf{How to construct a graph in traffic signal control}?
In real-world scenarios, intersections often exhibit complicated structures. 
For instance, different roads can be connected to the same intersections, each with a varying number of lanes.
To address this complexity, we can create a directed traffic light adjacency graph that reflects the spatial relationships among traffic lights. 
In this graph, nodes represent intersections, with additional states indicating the presence of traffic lights. The edges represent the roads connecting these intersections, with directional attributes specifying whether roads are one-way or two-way.
Moreover, additional features, such as the number of lanes on each road, the queue length and the number of vehicles in the lane, and the average speed of vehicles, should be considered. 
Models' performance can be assessed based on the average travel time of vehicles \cite{wang2020stmarl}. 

\textbf{How standard GNNs are modified for traffic signal control?}
In the context of traffic signal control, the adaptation of standard GNNs involves addressing the unique challenges of dynamic traffic environments and the interdependencies between intersections.
The first adaptation involves integrating GNNs with multi-agent reinforcement learning (RL) to effectively manage multiple intersections. Each intersection is controlled by an RL agent operating collaboratively within a Markov Decision Process. By employing multi-agent RL, GNNs enable these agents to model the spatial relationships and geographical structures of intersections, facilitating seamless information sharing and integration.
The second adaptation involves combining GNNs with components that process sequential data, such as RNNs or LSTM networks. This approach captures temporal patterns by using GCNs or GATs to understand spatial relationships, while LSTMs handle the temporal sequencing of traffic data, which is crucial for predicting future states.
Furthermore, the incorporation of attention mechanisms allows for differential weighting of the influence of various intersections and temporal phases.
These adaptations render GNNs as a dynamic solution capable of handling the temporal and spatial complexities inherent in traffic signal control systems.

\subsection{Transportation Safety}
Transportation safety aims to identify high-risk areas for accidents and incidents, crucial for pinpointing hotspots for future improvements. 
By understanding the factors contributing to these risks, researchers can develop targeted interventions and policies to mitigate them, thereby enhancing overall transportation system safety.
\begin{table*}[]
\centering
\caption{A Comprehensive Overview of Most Related Studies for Transportation Safety}
\label{table:transporation_safety}
\resizebox{\textwidth}{!}{%
\begin{tabular}{ccclcccl}
\hline
  \multicolumn{1}{c}{\textbf{Model}}&
  \multicolumn{1}{l}{\textbf{Article}} &
  \multicolumn{1}{c}{\textbf{Year}} &
  \multicolumn{1}{c}{\textbf{Datasets}} &
  \multicolumn{1}{c}{\textbf{Spatial Granularity}} &
  \multicolumn{1}{c}{\textbf{Temporal Granularity}} &
  \multicolumn{1}{c}{\textbf{Solution of zero-inflated}} &
  \multicolumn{1}{c}{\textbf{Summary}} \\ \hline
MSGNN &
  \cite{tran2023msgnn} &
  \multicolumn{1}{l}{2023} &
  \begin{tabular}[c]{@{}l@{}}loop detectors, \\ GPS probe data\\ (Brisbane, Gold Coast)\end{tabular} &
  Region Level &
  Short-Term (1h) &
  clustering-based data imputation &
  \begin{tabular}[c]{@{}l@{}} MSGNN, as a sub-area level accident prediction model, captures spatio and \\ temporal relations and uses a data imputation approach for sparse datasets.\end{tabular} \\ \hline
TAP &
  \cite{liu2023tap} &
  2023 &
  \begin{tabular}[c]{@{}l@{}}Real world dataset\end{tabular} &
  Region Level &
  Short-Term (30min) &
  - &
  \begin{tabular}[c]{@{}l@{}} Multi-task learning framework (TAP)  predicts traffic accidents using Spatio-\\ temporal Variational Graph Auto-Encoders, accelerated by edge computing\end{tabular} \\ \hline
GGCMT &
  \cite{huang2022deep} &
  \multicolumn{1}{l}{2022} &
  \begin{tabular}[c]{@{}l@{}}Real world dataset \\ (NYC) \end{tabular}&
  Region Level &
  Mid-Term (3h) &
  prior risk data enhancement method. &
  \begin{tabular}[c]{@{}l@{}} Computer-vision-based GGCMT predicts accidents using a gated graph convol-\\utional multi-task model, improved with prior risk data enhancement methods.\end{tabular} \\ \hline
DSTGCN &
  \cite{yu2021deep} &
  2021 &
  Real world dataset &
  Link Level &
  Short-Term &
  Negative sample undersampling method &
  \begin{tabular}[c]{@{}l@{}} DSTGCN predicts traffic accidents from heterogeneous data, capturing spatio-\\ temporal correlations by Deep Spatio-Temporal Graph Convolutional Network.\end{tabular} \\ \hline
GSNet &
  \cite{wang2021gsnet} &
  2021 &
  \begin{tabular}[c]{@{}l@{}} Real world dataset \\ (NYC, Chicago) \end{tabular} &
  Region Level &
  Short-term(1h) &
  Weight loss function &
  \begin{tabular}[c]{@{}l@{}} GSNet predicts traffic accidents by analyzing spatio-temporal and \\ heterogeneous data, using a specialized loss function for rare events.\end{tabular} \\ \hline
GraphCast &
  \cite{zhang2020multi} &
  \multicolumn{1}{l}{2020} &
  \begin{tabular}[c]{@{}l@{}} Real world dataset \\ (NYC) \end{tabular} &
  Region Level &
  Long-term (7day) &
  - &
  \begin{tabular}[c]{@{}l@{}} GraphCast, a multi-modal graph neural network, forecasts fine-grained \\ traffic risks in cities by combining social media and remote sensing data. \end{tabular} \\ \hline
RiskSeq &
  \cite{zhou2020foresee} &
  2020 &
  \begin{tabular}[c]{@{}l@{}} Real world dataset \\ (NYC, Suzhou) \end{tabular} &
  Region Level &
  Short-term (10min) &
  priori knowledge-based data enhancement &
  \begin{tabular}[c]{@{}l@{}} RiskSeq, a fine-grained and multi-step accident prediction model, combi-\\nes Differential Time-varying GCN and hierarchical sequence learning. \end{tabular} \\ \hline
RiskOracle &
  \cite{zhou2020riskoracle} &
  2020 &
  \begin{tabular}[c]{@{}l@{}} Real world dataset \\ (NYC, Suzhou) \end{tabular}&
  Region Level &
  Short-term (30min) &
  priori knowledge-based data enhancement &
  \begin{tabular}[c]{@{}l@{}} RiskOracle, a minute-level accident prediction model, combines Differential \\ Time-varying Graph network, multi-task and region selection strategies. \end{tabular} \\ \hline
TA-STAN &
  \cite{zhu2019ta} &
  2019 &
  \begin{tabular}[c]{@{}l@{}} Real world dataset \\(NYC) \end{tabular}&
  Region Level &
  Mid-Term (12h) &
  - &
  \begin{tabular}[c]{@{}l@{}} TA-STAN predicts accidents by analyzing real-world traffic data, vehicle \\ types, and external factors with a Spatial-Temporal Attention Network.\end{tabular} \\ \hline
\end{tabular}%
}
\vspace{-0.15in}
\end{table*}

\subsubsection{Spatial and Temporal Granularity}
In transportation safety, prediction models can be classified into four categories based on temporal and spatial granularity. Prediction periods divide models into long-term (day-level) \cite{yuan2018hetero, huang2019deep, zhang2020multi} and mid-term (hour-level) \cite{bao2019spatiotemporal, chen2018sdcae, ren2018deep, zhou2020riskoracle, huang2022deep}. Prediction regions distinguish models into link-level \cite{yu2021deep, zhu2019ta} and region-level \cite{tran2023msgnn, wang2021gsnet, wang2021spatio}.

Zhou et al. ~\cite{zhou2020riskoracle} introduced a three-stage RiskOracle framework for minute-level citywide traffic accident prediction using a Multi-task Differential Time-varying GCN (Multi-task DTGN) to model dynamic subregion correlations. 
Zhang et al. ~\cite{zhang2020multi} proposed GraphCast, a multi-modal sensing and GNN-based approach for regional accident prediction, utilizing social media and remote sensing data to handle noisy and heterogeneous multi-modal data. 
Tran et al. ~\cite{tran2023msgnn} developed a Multi-structured GNN (MSGNN) for predicting traffic incidents across entire networks, extracting area-wide features from various data sources for faster and more efficient incident prediction rather than link-level synchronization and map-matching. 
Huang et al. \cite{huang2022traffic} noted that many machine learning techniques predict the number of traffic accidents in each cell of a discretized grid without considering the underlying graph structure of road networks. Accurate link-level prediction requires a complex fusion of heterogeneous data resources and "map-matching" to represent all data with different granularity in the same map system.

\subsubsection{Spatial-temporal Correlation}
The work by Yu et al. \cite{yu2021deep} and Yuan et al. \cite{yuan2018hetero} both highlight that existing methods either ignore spatial-temporal correlations or make predictions at a coarse-grained level without considering the underlying graph structure of road networks.

Zhou et al.\cite{zhou2020foresee} proposed RiskSeq that addresses sporadic events with a self-adaptive ranking method. It employs a Differential Time-varying GCN (DT-GCN) enhanced with node-wise proximity and signal-wise differential operations to capture dynamic traffic and accident correlations. The framework also includes a Context-Guided LSTM for decoding risks across multiple spatial scales, focusing on spatiotemporal multi-granularity urban traffic risk prediction.
Yu et al. \cite{yu2021deep} tackled link-level accident prediction with a spatio-temporal convolutional network framework. Their model predicts link-level incident risk by learning spatial-temporal features from a road network graph, using graph convolutional operations to capture dynamic spatial and temporal variations.
Liu et al. \cite{liu2023tap} proposed a multi-task learning framework (TAP) based on edge computing, using spatio-temporal variational graph auto-encoders to enhance prediction accuracy by analyzing dynamic spatial-temporal traffic data correlations and integrating external factors.
Wang et al. \cite{wang2021gsnet} introduced GSNet, a region-wide accident prediction model capturing geographical and semantic spatial-temporal correlations. The model features a weighted loss function to address the zero-inflation issue.
Table.\ref{table:transporation_safety} lists other spatial-temporal models \cite{zhu2019ta, zhang2020graph}.

\subsubsection{Discussion}
\textbf{How to construct a graph in transportation safety?}
A straightforward way to construct a graph in transportation safety is using zone-based methods, which models target area as several medium-sized regions or grids, represented as a vertex in the graph. 
Edges between vertices indicate the connectedness between subregions, existing when traffic elements within those subregions have strong correlations. 
More specifically, the traffic elements contributing to the vertex features include static road network features (e.g., road types and number of lanes) and dynamic traffic features (e.g., traffic volume and average speed). These elements help define the relationships or connectedness within the urban graph.

\textbf{How standard GNNs are modified for transportation safety?}
When adapting GNN to transportation safety, a major challenge is dealing with the zero-inflated problem due to the rarity of accidents.
As we create zone-based graphs for GNN, the frequency of zero values in the training data increases with higher spatio-temporal resolution, making it harder to make accurate predictions.
Existing work addresses imbalanced anomaly data through various methods, including loss function handling \cite{huang2022deep, wang2021gsnet} and data preprocessing techniques like priori knowledge-based data enhancement \cite{zhou2020riskoracle, zhou2020foresee}, negative sample undersampling \cite{yu2021deep}, and graph augmentation \cite{wang2022contrastive}. 
Wang et al. \cite{wang2022contrastive} employed graph augmentation and contrastive loss to enhance latent representations in training, proposing an improved contrastive GNN-based framework for traffic anomaly analysis in ITS. Yu et al. \cite{yu2021deep} used negative sample undersampling to balance data by matching non-accident samples with accident samples, leveraging spatial-temporal GNNs for more accurate predictions. Wang et al. \cite{wang2021spatio} explored a chain-like triggering mechanism connecting accident occurrences, utilizing Spatial-Temporal Categorical GNNs (STC-GNN) to manage multi-dimensional and chain effects for fine-grained temporal accident prediction.


\subsection{Traffic Demand Forecasting}

Traffic demand prediction involves predicting the volume of users needing transport to or from specific areas or locations, which is crucial for scheduling services efficiently and enhancing overall transport system responsiveness.  
However, the growth of modern cities has caused an increase in traffic-related issues, which puts much pressure on public transportation systems. 
To tackle this problem, ride-hailing services such as Uber, Lyft, and DiDi, as well as bike-sharing services like MoBike have emerged as potential solutions \cite{tong2017simpler,zi2021tagcn}. 
As a result, there is now a pressing need for accurate traffic demand prediction that can precisely forecast future crowd demands, thus scheduling future transportation services and other downstream tasks.\cite{toole2015path}. 
However, predicting traffic demand poses challenges.
On the one hand, human behavior, influenced by weather and special events, introduces variability that makes accurate forecasting even more complicated. 
On the other hand, integrating diverse data sources for comprehensive analysis requires advanced models that are capable of processing complex, large-scale datasets in real-time.

\subsubsection{Traffic Zone-based Graph Methods}
One of the pioneering works in utilizing graph learning methods for demand prediction tasks is Spatio-Temporal Multi-Graph Convolution Network (ST-MGCN) \cite{geng2019spatiotemporal}. It proposes to exploit graph structures from multiple perspectives to capture comprehensive information on the spatio-temporal characteristics of traffic systems. 
Specifically, ST-MGCN builds the graphs of zones from three angles: a neighborhood graph based on the spatial proximity, a functionality graph defined by the POI similarity, and a transportation connectivity graph induced by road networks such as motorways, highways, or public transportation systems like subways. 
A multi-graph convolution is then introduced to model the spatial dependencies between regions and provide informative representations for downstream demand prediction tasks. 
Similarly, PGDRT \cite{lee2023pgdrt} builds the zone-wise relational graph using three types of temporal characteristics: adjacent visual characteristics, periodic characteristics, and representative characteristics, to provide a more comprehensive view of temporal features in traffic systems. 
To fully exploit the rich information from multiple traffic systems, Multiview Spatio-Temporal Graph Neural Networks (MSTGNN) \cite{zhao2023developing} proposes a multiview graph that jointly depicts the demand relationship between bus, metro, and taxi demands. The multiview graph enables MSTGNN to capture the interaction dependencies among the travel demands of different transportation systems. An auxiliary loss is used to encourage the consistency between graph features from multiple views and enhance the performance of TGCN modules.

\subsubsection{Spatio-temporal Graph-based Methods}
Classical GNN models for traffic demand prediction treat the spatial dependency as a static graph and cannot depict dynamic features.
However, in reality, the spatial dependencies between most nodes change over time, while others remain relatively constant.
To address this limitation, the Dynamical Spatio-Temporal Graph Neural Network (DSTGNN) \cite{huang2022dynamical} evaluates the stability of a node's spatial dependence based on the number of dissimilar neighbors and constructs a spatio-temporal graph that evolves over time. 
To encode the spatio-temporal information, the model uses a spatio-temporal embedding network that combines a Diffusion Convolution Neural Network (DCNN) with a modified transformer.

\begin{table*}[]
\centering
\caption{A Comprehensive Overview of Most Related Studies for Demand Prediction}
\label{table:demand_prediction}
\resizebox{\textwidth}{!}{%
\begin{tabular}{cccccccl}
\hline
  \multicolumn{1}{c}{\textbf{Model}} &
  \multicolumn{1}{c}{\textbf{Article}} &
  \multicolumn{1}{c}{\textbf{Year}} &
  \multicolumn{1}{c}{\textbf{Prediction Task}} &
  \multicolumn{1}{c}{\textbf{Graph Views}} &
  \multicolumn{1}{c}{\textbf{GNN Module}} &
  \multicolumn{1}{c}{\textbf{Temporal Module}} &
  \multicolumn{1}{c}{\textbf{Summary}} \\ \hline
  PGDRT & \cite{lee2023pgdrt} & 2023 & Taxi Passenger & Neighborhood, Function, Connectivity & GCN & ConvLSTM & \begin{tabular}[c]{@{}l@{}} PGDRT considers a region’s unique characteristics and the influence of \\ regions on the model of the dependent relationship between regions.\end{tabular} \\
  \hline
  MSTGNN &\cite{zhao2023developing} & 2023 & Bus, Metro, Taxi & Neighborhood, Connectivity & GCN & Temporal GCN & \begin{tabular}[c]{@{}l@{}} MSTGNN uses a multiview graph consisting of bus, metro, and taxi \\ views, with each view containing both local and global graphs.\end{tabular}\\
  \hline
  STGMT & \cite{wen2023traffic} & 2023 & Taxi \& Highway & Traffic Network & Node2Vec & Multi-head Attention & \begin{tabular}[c]{@{}l@{}} 
  STGMT combines Multi-head Temporal Attention (MTA) and Multi-\\head  Temporal Interactive Attention (MTIA) for temporal features.
  \end{tabular}\\
  \hline
  PAG-TSN & \cite{li2023pag} & 2023 & Ride-hailing & Distance, POI relation & BAT-GCN & PA-GRU & \begin{tabular}[c]{@{}l@{}} PAG-TSN uses a bicomponent attention GCN and a periodic attent-\\ional GRU to integrate the extracted spatio-temporal information. \end{tabular} \\
  \hline
  HetGNN-LSTM & \cite{nazzal2023semi} & 2023 & Taxi & Decentralized taxi graph & HetGNN & LSTM & \begin{tabular}[c]{@{}l@{}} HetGNN-LSTM proposes a semi-decentralized approach utilizing \\ multiple cloudlets, moderately sized storage, and computation devices.\end{tabular}\\
  \hline
  MFGCN & \cite{liao2023mfgcn} & 2023 & Ride-hailing & OD network & MODGCN & TAS-LSTM  & \begin{tabular}[c]{@{}l@{}} MFGCN is a multimodal fusion GCN that consists of a multimodal \\ module to incorporate weather and temporal activity patterns. \end{tabular} \\
  \hline
  SGCNPM & \cite{yang2023short} & 2023 & Dockless Bike-Sharing & Distance, Function, Interconnectio & MGCN & LSTM & \begin{tabular}[c]{@{}l@{}} SGCNPM considers time, built environment, and weather to create \\ a prediction method considering the influence of multiple factors.\end{tabular}\\
  \hline
  DSTGNN & \cite{huang2022dynamical} & 2022 & Taxi \& Bike & Spatial dependency & DCNN & Multi-head Attention & \begin{tabular}[c]{@{}l@{}} DSTGNN builds spatial graphs based on the stability of the node’s \\ spatial dependence to capture the dynamical relationship.\end{tabular}\\
  \hline
  DMVST-VGNN & \cite{jin2022deep} & 2022 & Ride-hailing & Multi-view Graph Generation & GAT & Multi-head Attention & \begin{tabular}[c]{@{}l@{}} The Model integrates 1D CNN, Multi-Graph Attention Neural Networks, \\and Transformer to construct multiview spatio-temporal information.\end{tabular}\\
  \hline
  ST-MGCN & \cite{geng2019spatiotemporal} & 2019 &  Ride-hailing & Neighborhood, Function, Connectivity & ChebNet & RNN & \begin{tabular}[c]{@{}l@{}} ST-MGCN uses GNNs to model non-Euclidean pair-wise correlations \\ between different regions by designing a spatio-temporal multi-graph. \end{tabular}  \\
  \hline
\end{tabular}%
}
\vspace{-0.2in}
\end{table*}

\subsubsection{Dynamic Graph-based Methods}
The traditional approach to modeling cities is to divide them into grid-like zones and construct graphs based on these divisions. 
However, this approach can lead to suboptimal solutions, and adapting to dynamic graph structures remains challenging.
A new solution called Deep Multi-View Spatio-temporal Virtual Graph Neural Network (DMVST-VGNN) \cite{jin2022deep} improves learning capabilities related to spatial dynamics and long-term temporal dependencies. 
The DMVST-VGNN method proposes a graph generation process that provides a more flexible and fine-grained perspective on the spatio-temporal relationships between regions, as opposed to the simplistic grid-based division of the map. 
Another proposal by Nazzal et al. \cite{nazzal2023semi} extends the idea of dynamic and flexible graph structures to decentralized edge-computing scenarios and introduces a heterogeneous GNN-LSTM algorithm. 
This algorithm is designed to handle dynamic taxi graphs where taxis serve as nodes. The proposed heterogeneous GNN-LSTM structure has demonstrated the ability to capture dynamic decentralized graph structures and has shown promising results in taxi-level demand and supply forecasting.

\subsubsection{Improvements on Graph Encoders}
The traditional graph convolution network has limited capability to represent the complex information in traffic zone graphs. 
However, some works aim to enhance the expressiveness of graph encoders. 
STGMT \cite{wen2023traffic} proposes the Sandwich-Transformer for processing spatio-temporal traffic graphs, which is composed of a Multi-head Temporal Attention (MTA) and a Multi-head Temporal Interactive Attention (MTIA). 
PAG-TSN \cite{li2023pag} constructs a Bicomponent Attention Graph Convolution model (BAT-GCN) and a periodic attentional gated recurrent unit model to capture geographical relationships and temporal features of different periods, respectively. 
While previous research primarily concentrates on processing plain time-series traffic demand data for predictions, it is essential to recognize that contextual information and multimodal attributes, such as weather conditions, significantly impact ride-hailing and other public traffic systems. 
To tackle these challenges, Multimodal Fusion Graph Convolutional Network (MFGCN) \cite{liao2023mfgcn} introduces an innovative Multimodal Fusion Graph Convolutional Network for traffic demand prediction. MFGCN incorporates a Multimodal Origin-Destination GCN (MODGCN) that comprises three GCNs to capture spatial patterns and a Multimodal Attribute Enhancement (MAE) module for integrating dynamic weather and metadata.
SGCNPM \cite{yang2023short} utilizes multiple modules that consist of GCN and LSTM operators to model the multiple factors in a dynamic traffic system, including time periods, built environment, and weather, to predict the short-term demand of a dockless bike-sharing system. A comprehensive overview of most related studies for demand prediction can be found in Table.\ref{table:demand_prediction}.

\subsubsection{Discussion}

\textbf{How to construct a graph in demand prediction?}
In demand prediction, in addition to the zone-based construction approach discussed in the transportation safety section, we can also create a traffic graph based on network topology, where nodes represent stations, bus stops, etc., and edges can be constructed based on the correlation among the features of the nodes.
Research has shown the effectiveness of location-based graphs, but constructing graphs from multiple perspectives has proven to introduce more informative relationships between geographical zones. 
For instance, the Spatio-Temporal Multi-Graph Convolution Network (ST-MGCN) \cite{geng2019spatiotemporal} builds zone graphs from three perspectives: a neighborhood graph based on geographical distances between neighborhoods, a functionality graph derived from the similarity between POI vectors, and a transportation connectivity graph induced by road networks such as motorways, highways, or public transportation systems like subways.

\textbf{How standard GNNs are modified for demand prediction?}
In the field of demand prediction, it is essential to adapt GNNs to accommodate the spatio-temporal characteristics of the input flow graph. This involves incorporating temporal and spatial modules and other necessary modifications. 
Additionally, based on the enhancement of the multi-view approach we have explored \cite{lee2023pgdrt,zhao2023developing,jin2022deep}, a multi-view fusion mechanism is required to aggregate from different representations.  
Furthermore, to capture the dynamic nature of demands, GNNs can be combined with other sequence models such as RNNs \cite{nazzal2023semi,lee2023pgdrt,li2023pag} and self-attention layers \cite{wen2023traffic,huang2022dynamical}.


\subsection{Parking Management}
Parking management has become a significant concern due to the limited resources of parking spaces and the increasing traffic pressure.
One crucial research area of intelligent parking management is predicting parking availability, which involves forecasting future parking occupancy. 
Being able to predict parking availability on a city-wide scale is crucial for developing effective Parking Guidance and Information (PGI) systems, such as Baidu Map  \cite{rong2018parking} and Google Map \cite{arora2019hard}. 

Predicting the availability of parking spaces faces several challenges, such as the non-Euclidean spatial autocorrelation between parking lots, the dynamic temporal autocorrelation within and between parking lots, and the information scarcity due to the lack of real-time data obtained from sensors to determine parking availability.
GNNs have been recognized as an effective approach for processing spatial-temporal and graph-based structures, addressing the mentioned challenges, and providing more accurate prediction accuracy.

As one of the pioneering works on modeling the parking availability prediction with graph-based models, SHARE \cite{zhang2020semi1} and its variant SHARE-X \cite{zhang2020semi} proposes a Semi-supervised Hierarchical Recurrent Graph Neural Network to analyze spatio-temporal parking data. 
Specifically, SHARE proposes a hierarchical graph convolution module that captures non-Euclidean spatial correlations between parking lots. 
It consists of two blocks: a contextual graph convolution block for local spatial dependencies and a soft clustering graph convolution block for global spatial dependencies. 
SHARE-X extends the idea of SHARE to address the lack of real-time sensors in real-world scenarios. 
Particularly, It leverages a parking availability approximation module to estimate parking availability for parking lots without sensor monitoring.

To better depict the strong spatiotemporal contextual autocorrelation between vacant parking spaces, dConvLSTM-DCN \cite{feng2022predicting} analyzed the historical zone-wise parking space data and found that there is both a temporal correlation within each parking lot and a spatial correlation among different parking spaces.
Based on this observation, Feng et al. proposed a deep learning framework called dual Convolutional Long Short-Term Memory with Dense Convolutional Network (dConvLSTM-DCN) to predict the availability of vacant parking places in the short-term (within 30 minutes) and long-term (over 30 minutes) zone-wisely.
The framework consists of two parallel ConvLSTM components that capture the spatial correlations among parking lots and provide an informative representation of the prediction process.

The traditional methods of obtaining real-time on-street parking information rely heavily on densely deployed sensors. 
Therefore, the high costs of current parking availability prediction models have made it difficult to use them in many cities and areas. 
To tackle this issue and avoid the high costs of deploying new sensors, MePark \cite{zhao2021mepark} aims to predict real-time on-street parking availability across a city by using existing infrastructure and easily accessible data without relying solely on specially deployed sensors. 
More specifically, MePark uses an iterative mechanism to combine the aggregated inflow and individual parking duration predictions to exploit the transaction data adequately.
Additionally, it extracts distinctive features from multiple data sources, combining the MGCN and the LSTM network to capture complex spatio-temporal correlations.

\subsubsection{Discussion}
\textbf{How to construct a graph in parking management?}
Constructing the graph in the parking management system follows a similar approach to previous research domains, such as transportation safety and demand prediction. 
The zone-based construction method can predict parking availability in a medium-sized grid or area, while the network topology-based method can predict availability in specific parking lots based on their geometric information.

\textbf{How standard GNNs are modified for parking management?}
Firstly, the dynamic temporal autocorrelation and non-Euclidean relationship between parking lots is crucial for accurate short-term parking availability prediction. 
Many existing efforts suggest integrating recurrent network structures, such as LSTMs, into GNN architectures \cite{zhang2020semi1, feng2022predicting, zhao2021mepark}, while some methods directly reform GNNs with recurrent structures, such as SHARE and SHARE-X \cite{zhang2020semi}.
Secondly, information scarcity poses another challenge, prompting researchers to explore GNNs to exploit non-sensor data, like transaction data, to harness the prediction potential \cite{zhao2021mepark}.

\section{Challenges and Future Directions}
\label{challenges}
After analyzing the current studies on GNNs in ITS, we discuss the challenges and future directions for applying GNNs in ITS. This is important to identify any gaps that need to be addressed and to provide insights for further research.

\subsection{Research Challenges}


\subsubsection{Data}

Constructing datasets is one of the primary challenges when applying GNNs in ITS. 
Data privacy is a major concern when collecting information from traffic sensors or GPS data. 
Currently, there are limited publicly available data sources, such as Data.gov, The University of Sydney’s Intelligent Vehicles and Safety Systems, and the Connected Vehicle DataSets from the Safety Pilot Model Deployment \cite{guerrero2021deep}. Some researchers have explored multi-modal models to access data from richer sources like social media \cite{liu2023multi, fafoutellis2023traffic, wang2023routing, liao2023mfgcn}. However, these sources often face issues related to credibility and a lack of valuable information.
As a result, generating a large, high-quality, and comprehensive dataset for ITS remains a formidable task, requiring continued effort to address data privacy concerns and improve data sources.
\subsubsection{Model}

\textbf{Domain-specific Model Design}.
The transportation network is a complex system that includes various nodes and edges, such as roads, intersections, and vehicles. Designing GNN models in ITS that can efficiently learn from such a heterogeneous and complex structure requires significant effort.
The design of GNN applications in ITS heavily depends on the specific goals of the corresponding applications, as different goals require using different graph models and construction techniques. For instance, GNNs are commonly used in traffic forecasting and travel demand modeling to predict features or variables over graph nodes. While in areas such as traffic signal control, GNNs focus on learning control policies or unraveling agent interactions that involve learning or predicting over edges or the entire graph.
Besides, GNNs face different challenges in various transportation domains. The pure GNN models can not effectively solve the problem, so some scholars have explored the potential of combining GNNs with other approaches. 
For instance, in decision-making problems, such as traffic signal control, reinforcement learning is an effective technique. When multiple intersections interact, multi-agent reinforcement learning methods combined with GNNs have been proposed \cite{nishi2018traffic, zeng2021graphlight,zeng2021graphlight}. 
In some particular scenarios, such as traffic accident prediction, positive samples like accidents can be rare when predicting within a fine-grained granularity. To improve accuracy, we can use data augmentation techniques like a priori knowledge-based data enhancement \cite{zhou2020riskoracle, zhou2020foresee} and negative sample undersampling methods \cite{yu2021deep}.
Nearly every transportation domain has its own domain-specific problems and unique characteristics. Therefore, combining GNNs and other techniques requires nuanced graph construction, tailored problem analysis, and painstaking design.

\textbf{Dynamic Spatio-temporal Dependency}. 
Utilizing GNNs to model spatio-temporal dependencies in ITS presents a significant challenge \cite{bui2022spatial}. This difficulty arises from the need to precisely capture both the complex and dynamic spatial interactions within the transportation network and the temporal dynamics that reflect the constantly changing traffic patterns.
Transportation networks frequently exhibit dynamic spatial dependencies, as the graph structure can evolve over time due to the constantly changing urban environment. For example, in trajectory prediction, it is crucial to identify key agents and objects, such as vehicles, cyclists, and pedestrians, that can influence the predicted trajectory. Consequently, a graph framework capable of adapting to these real-time changes is necessary to maintain accurate predictions.
Meanwhile, in terms of temporal dependencies, traffic conditions at any given moment are shaped by a multitude of past events. 
Accurately capturing these long-term dependencies is crucial for precise forecasting, but it poses significant computational challenges and demands sophisticated memory mechanisms within the model \cite{shao2022long, roy2021sst}. 
Furthermore, real-time data processing is vital for practical applications in ITS, adding another layer of complexity \cite{sharma2023graph, conlan2023real}. 
The model must integrate and process this multifaceted data while adapting in an environment marked by continuous change and uncertainty. Therefore, effectively modeling spatio-temporal dependencies remains a pivotal yet challenging aspect of utilizing GNNs in ITS.

\textbf{Robustness, Reliability, Interpretability}. 
Deep learning has faced criticism for its lack of interpretability and black-box nature, which makes it difficult to assess the rationale behind recommendations made by graph-based deep learning approaches in transportation safety and other high-stakes fields. 
Furthermore, it is essential to ensure that GNN models remain reliable in large-scale real-world applications, such as during rush hours, sensor failures, or cyber-attacks \cite{jiang2023enhancing}. 
As we strive to enhance model performance, we must also be vigilant about potential failures and undetected anomalies. Additionally, scalability remains a critical concern. Current GNN frameworks, including those based on TensorFlow, PyTorch, DGL, and PyG, face limitations in scalability, which restrict their application to large-scale graphs due to inadequate system support \cite{li2023graph}.


\subsubsection{Computation}
Processing, storing, and transmitting vast amounts of data has become increasingly critical in today’s extensive data landscape, particularly within ITS. While GNNs and deep learning techniques are extensively utilized in ITS, they face substantial challenges due to their high computational demands. These challenges are amplified when performing real-time or near-real-time inference and processing large volumes of data from extensive camera networks. Additionally, the limited resources of IoT devices—such as constrained memory and computing power—complicate these issues further. To address these challenges, researchers have proposed various solutions, including edge computing, graph sampling, hardware acceleration, and optimized algorithms.

\subsection{Future Directions}
\subsubsection{More Integration of Advanced Techniques}
As noted, GNNs are highly effective for capturing spatial-temporal relationships and making inferences on graph-based data structures. 
However, since different problems have unique characteristics and challenges, it is crucial to design tailored models for each specific issue. Additionally, integrating other techniques into GNN frameworks can significantly enhance model performance and facilitate practical applications.
For example, incorporating the edge learning paradigm \cite{zhang2021edge} into GNN frameworks can address the storage, memory, and computational limitations of data-producing devices. 
This approach allows distributed edge devices to collaboratively train models and perform inferences, thus ensuring privacy and security \cite{zhang2021edge}. 
Similarly, leveraging transfer learning \cite{mao2021transfer} and meta-learning \cite{wang2022meta} can significantly improve a model's adaptability to different cities with varying traffic patterns.
In conclusion, combining GNNs with advanced techniques such as reinforcement learning, transfer learning, meta-learning, generative adversarial networks (GANs), semi-supervised learning, and Bayesian networks opens new avenues for addressing domain-specific problems and challenges. This synergistic approach not only leads to more robust and versatile solutions but also unveils exciting possibilities for solving complex, real-world problems across various domains. As research continues, it is crucial to explore these integrations further, pushing the boundaries of what can be achieved with GNNs and their allied technologies.

\subsubsection{More Expanding Applications of GNNs}
Currently, most research on GNNs in ITS focuses on traffic prediction. However, significant untapped potential remains for further developing GNNs and exploring their broader applications within this field.
To enhance GNNs, efforts should be directed towards improving their efficiency, robustness, and generality. One approach is to implement multi-modal learning \cite{liu2023multi, yin2021multimodal}, which allows models to integrate a richer set of contextual information. Additionally, incorporating more complex graph structures, such as heterogeneous graphs \cite{nazzal2023semi} and hypergraphs \cite{yadati2019hypergcn}, can facilitate handling larger and more intricate graph structures.
Furthermore, exploring GNN applications in other ITS research domains also holds promise. For example, the 3D structural understanding required for autonomous vehicles presents a unique opportunity for GNNs, especially compared to traditional transformer architectures \cite{guo2021pct, zhao2021point} used in point cloud processing, which often lack efficiency.
By combining graph convolution with self-attention mechanisms, we can significantly enhance feature extraction and capture both local and global contexts effectively. This integration has the potential to greatly improve GNN performance in ITS applications \cite{lu20223dctn}.
While we have covered various domains, from traffic prediction to traffic safety, additional areas remain to explore, such as route planning, urban land-use planning, and traffic pattern recognition. Further investigation into these applications of GNNs could yield new insights and enhance their performance across a broader range of ITS scenarios.

\subsubsection{More Comprehensive Experiments}
Currently, some research experiments in the field of ITS rely on simulators. However, data generated by traffic simulation software may not always accurately reflect real-world conditions due to factors such as variations in driver behavior and alternative route planning \cite{zhancheng2021research}. 
Furthermore, even when models are evaluated using real-world data, such testing is often limited in scale, or the model's running time may not be fully reported. These limitations do not necessarily ensure the model's reliability, robustness, or generalization ability to real-world scenarios. 
According to work by Shi et al. ~\cite{shi2023improving}, some models, such as those based on Deep Q-Networks (DQNs) for reinforcement learning, experience performance degradation when dealing with large-scale road networks or incomplete data, complicating their generalization. 
Therefore, it is essential to conduct more comprehensive experiments using large-scale real-world data to fully assess and validate the effectiveness of these models.


\section{Conclusion} \label{conclusion}
With the rapid advancement of deep learning, GNNs have emerged as a promising tool in ITS. However, most existing research on GNNs in ITS has concentrated on traffic forecasting while overlooking other critical areas, such as autonomous vehicles and transportation safety.
In this work, we have reviewed and analyzed representative papers from 2018 to 2023 that explore various applications of GNNs across six domains within ITS. 
We have summarized and categorized these studies based on their research focus, the graph methods employed, and the domain-specific challenges encountered, and presented the findings in detailed tables and lists.
Our analysis reveals that while many studies are limited to specific functionalities of GNNs, such as modeling graph-structured data and capturing spatio-temporal relationships, a vast potential is waiting to be fully realized. 
GNNs can be expanded to other areas of ITS, opening up new possibilities and avenues for research and development.
Additionally, we have identified common challenges that need to be addressed when applying GNNs in ITS, including issues related to $data$, $model$, and $computation$. 
We also highlight future directions for GNN research in ITS, emphasizing the crucial role of integrating GNNs with other techniques. This approach can lead to more comprehensive solutions, broadening the application scope of GNNs and fostering more extensive experiments.

\IEEEdisplaynontitleabstractindextext

%
\IEEEpeerreviewmaketitle

\ifCLASSOPTIONcompsoc
  \section*{Acknowledgments}
\else
  \section*{Acknowledgment}
\fi


The authors are grateful to the anonymous reviewers for critically reading this article and for giving important suggestions to improve this article. 

This paper is partially supported by the National Key Research and Development Program of China with Grant No. 2023YFC3341203, and the National Natural Science Foundation of China (NSFC Grant Numbers 62306014 and 62276002).

\ifCLASSOPTIONcaptionsoff
  \newpage
\fi



%
\bibliographystyle{IEEEtran}
\bibliography{rec}

\end{document}